%% file: main.tex
\documentclass[10pt,twocolumn,letterpaper]{article}

\usepackage[pagenumbers]{wacv} 

\input{preamble}

\definecolor{wacvblue}{rgb}{0.21,0.49,0.74}
\usepackage[pagebackref,breaklinks,colorlinks,allcolors=wacvblue]{hyperref}

\def\wacvPaperID{509} 
\def\confName{WACV}
\def\confYear{2026}

\title{CLIP-UP: CLIP-Based Unanswerable Problem Detection \\ for Visual Question Answering}

\author{Ben Vardi$^1$ \quad\quad\quad\quad Oron Nir$^{1,2}$ \quad\quad\quad\quad Ariel Shamir$^1$\\
$^1$Reichman University \quad\quad  $^2$Microsoft Corporation\\
{\tt\small \href{https://benvr.github.io/CLIP-UP}{https://benvr.github.io/CLIP-UP}}\\
}

\begin{document}
\crefname{appendix}{App.}{Apps.}
\maketitle
\input{sec/0_abstract}
\input{sec/1_intro}
\input{sec/2_related_work}
\input{sec/3_method}

\input{sec/4_experiments}
\input{sec/5_conclusions}
\input{sec/acknowledgements} 
{
    \small
    \bibliographystyle{ieeenat_fullname}
    \bibliography{main}
}
\input{sec/X_suppl}

\end{document}

%% file: preamble.tex
\usepackage{multirow} 
\usepackage[table]{xcolor} 

\usepackage{tikz} 
\usepackage{textcomp} 
\newcommand{\doublestraightquotes}[1]{\textquotesingle\textquotesingle#1\textquotesingle\textquotesingle}

\usepackage{makecell}   
\usepackage{pifont}     
\newcommand{\cmark}{{\color{green!60!black}\ding{51}}} 
\newcommand{\xmark}{{\color{red}\ding{55}}}            

\newcommand{\graymidrule}{\arrayrulecolor{lightgray}\midrule\arrayrulecolor{black}}

\newcolumntype{L}[1]{>{\raggedright\arraybackslash}p{#1}}

\makeatletter
\AddToHook{cmd/appendix/before}{\def\cref@section@alias{appendix}\def\cref@subsection@alias{appendix}}
\makeatother

%% file: sec/0_abstract.tex
\begin{abstract}

Vision-Language Models (VLMs) demonstrate remarkable capabilities in visual understanding and reasoning, such as in Visual Question Answering (VQA), where the model is asked a question related to a visual input. 
Still, these models can make distinctly unnatural errors, for example, providing (wrong) answers to unanswerable VQA questions, such as questions asking about objects that do not appear in the image.

To address this issue, we propose CLIP-UP: CLIP-based Unanswerable Problem detection, a novel lightweight method for equipping VLMs with the ability to withhold answers to unanswerable questions. 
CLIP-UP leverages CLIP-based similarity measures to extract question-image alignment information to detect unanswerability, requiring efficient training of only a few additional layers, while keeping the original VLMs' weights unchanged.

Tested across several models, CLIP-UP achieves significant improvements on benchmarks assessing unanswerability in both multiple-choice and open-ended VQA, surpassing other methods, while preserving original performance on other tasks. 

\end{abstract}

%% file: sec/1_intro.tex
\section{Introduction}
\label{sec:intro}

A fundamental task in the domain of Vision-Language Models (VLMs) is Visual Question Answering (VQA)~\cite{antol2015vqa}, where the model is asked a question related to a visual input. 
VQA appears in various formats, with open-ended questions being the most immediate
form.
Multiple-choice VQA offers a more constrained setup, requiring to discriminate between several plausible answer options.

Recent VLMs~\cite{liu2024llavanext,zhu2025internvl3exploringadvancedtraining,lu2024ovisstructuralembeddingalignment} excel in both VQA formats but a critical challenge persists: the tendency of VLMs to produce incorrect or irrelevant responses, a phenomenon commonly referred to as ``hallucinations''~\cite{rohrbach2018object, liu2024survey}. The problem is particularly concerning in VQA, where models might confidently provide answers that, while seemingly plausible, are actually inconsistent with the visual content~\cite{miyai2025unsolvable,li2023evaluating}. This raises concerns about VLMs' reliability and applicability, indicating the need for mitigating such hallucinations~\cite{bai2024hallucination}.

\begin{figure*}[t]
    \centering
    \includegraphics[width=0.89\linewidth]{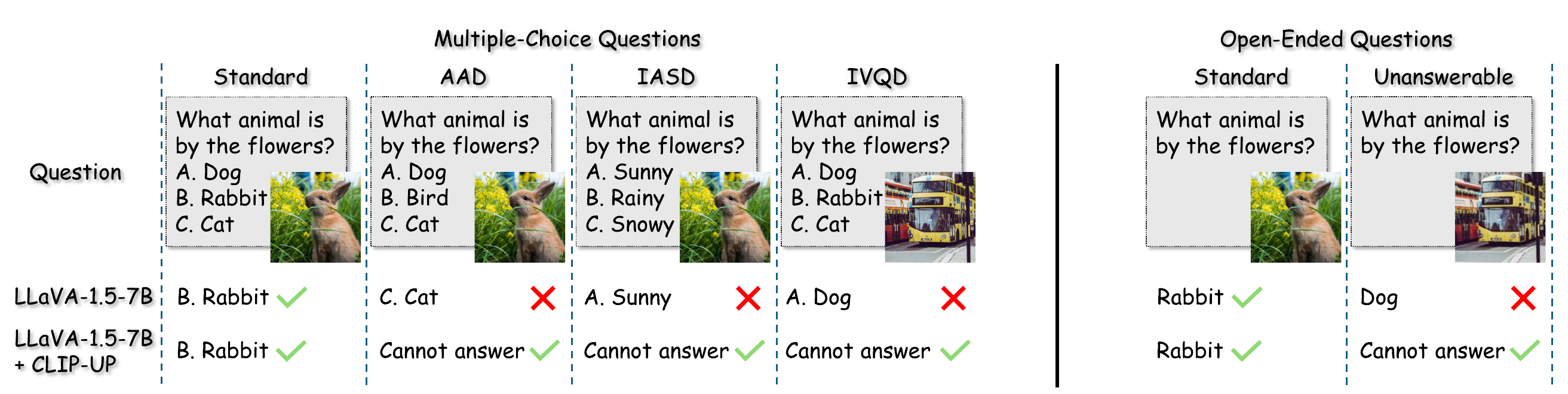}
  \caption{CLIP-UP equips VLMs such as LLaVA-1.5-7B~\cite{liu2023improvedllava} with the ability to detect and withhold answers to multiple-choice and open-ended unanswerable questions, while preserving models' original capabilities on standard answerable questions.}
  \label{fig:problem_demo}
\end{figure*}

In this work, we focus on a crucial hallucination challenge within VLMs: their tendency to provide answers to unanswerable visual questions, flawed either in the question-image pairing or the question itself, in both multiple-choice and open-ended VQA.
For multiple-choice VQA, this issue was formalized as the \emph{Unsolvable Problem Detection (UPD)} challenge~\cite{miyai2025unsolvable}.
In this challenge, models are evaluated first on their ability to detect unanswerable VQA inputs and withhold answers when necessary, and second on correctly answering answerable questions. Hence, this task is more complex than a simple binary classification of unanswerability. Visual unanswerable multiple-choice questions are classified into three categories~\cite{miyai2025unsolvable}:
(1) Absent Answer Detection (AAD): detecting questions where all answer options are incorrect; 
(2) Incompatible Answer Set Detection (IASD): detecting questions with answer options incompatible with the question; 
and (3) Incompatible Visual Question Detection (IVQD): detecting questions where the question is incompatible with the image (see \cref{fig:problem_demo} for illustration).

The importance of UPD in multiple-choice VQA lies not only in improving model robustness but also in the fact that multiple-choice questions are commonly used to benchmark VLMs, as the structured format simplifies evaluation compared to open-ended questions.
However, this format can obscure the actual understanding of the model.
For example, rather than genuinely reasoning about the input, models may rely on shortcut strategies, such as eliminating unlikely options. 
Equipping models with the ability to withhold answers to unanswerable questions discourages such behavior and enables more reliable evaluations that better reflect models' true understanding.

For open-ended VQA, the importance of UPD lies in ensuring that models are robust and capable of handling unanswerable questions that arise in real-world scenarios, such as when assisting visually impaired individuals~\cite{gurari2018vizwiz}.

An examination of VLMs' training data reveals a predominance of valid, answerable questions~\cite{liu2023improvedllava}, likely contributing to their tendency to always provide an answer.
A straightforward solution to UPD is therefore to fine-tune or re-train the full model with data that include unanswerable questions. However, this approach is often impractical due to the high computational cost and data requirements. 
Therefore, a pressing question arises: how can we adapt existing models to identify when they should refrain from answering in affordable costs?

To tackle this question, \citet{miyai2025unsolvable} and \citet{qian2024easyfoolmultimodalllms} explored prompt engineering solutions that modify the input to inform the model it may withhold an answer. 
This is done, for example, by adding a ``None of the above'' option in multiple-choice questions, providing an alternative when the question is unanswerable.
While appealing due to its training-free nature and ease of implementation, prompt engineering was found to provide only limited improvement. 
Another solution is to fine-tune models on VQA-specific data, including only answerable and unanswerable questions~\cite{miyai2025unsolvable, chandu2025certainlyuncertain}. Although this approach significantly improves models' ability to detect unanswerable questions, its exclusive focus on VQA compromises performance on tasks other than VQA.

In this work, we introduce CLIP-based Unanswerable Problem detection (CLIP-UP): a lightweight method for enhancing general pre-trained VLMs with the capability to detect unanswerable questions in both multiple-choice and open-ended VQA formats (see \cref{fig:problem_demo}).
CLIP-UP operates by leveraging carefully crafted \emph{correlation vectors} derived from CLIP embeddings~\cite{radford2021learning} of the input image and question, encoding image-question alignment information.
These vectors are projected into the VLM’s intermediate feature space, producing a new embedding vector that serves as an answerability prior and is seamlessly integrated into the model.
CLIP-UP this way trains only a single linear projection layer for each question format (multiple-choice or open-ended) to create this vector, while keeping the original VLM weights unchanged.

Using a simple classifier, we determine whether and which embedding vector should be activated: 
for multiple-choice or open-ended VQA inputs, the corresponding vector is generated and integrated to enhance UPD capabilities; 
otherwise, no vector is generated, ensuring the model’s original capabilities remain intact on non-VQA tasks.

The effectiveness of CLIP-UP depends on the quality of the CLIP signal and the use of this information.
In particular, we use Structure-CLIP~\cite{huang2023structure}, a CLIP variant that offers improved sensitivity for texts with similar structures but different semantics.
Beyond using correlation vectors to generate an embedding vector fed into the VLM, we also propose \emph{Injected LoRA}, a novel approach for injecting priors directly into LoRA~\cite{hu2021lora} layers, where the prior in our case is the correlation vectors.

Our experiments on multiple VQA unanswerability benchmarks show that the UPD problem persists in recent VLMs and that CLIP-UP, applied across different models, achieves performance surpassing other UPD enhancement methods.
We also created a new multiple-choice dataset for training CLIP-UP, covering all unanswerability categories.

Our contributions can be summarized as follows:
\begin{itemize}    
    \item We introduce CLIP-UP, a novel lightweight approach leveraging 
    CLIP to enhance VLMs' ability to withhold answers to unanswerable questions, while leaving the original VLM weights unaltered.
    \item We demonstrate that CLIP-UP significantly improves UPD performance in multiple-choice and open-ended VQA across various models, outperforming other methods while preserving performance on non-VQA tasks.
    \item We propose a new method for injecting priors into LoRA layers, and show that this way of using CLIP-UP correlation vectors improves results over standard LoRA.
    \item We release our code and multiple-choice UPD training dataset to support future research.
\end{itemize}

%% file: sec/2_related_work.tex
\section{Related work}
\label{sec:related_work}

\paragraph{Vision-language models}

The rapid advancements in Large Language Models (LLMs) in recent years~\cite{touvron2023llama,chiang2023vicuna,chowdhery2023palm,brown2020gpt3} have led to impressive performance across a wide range of text-based tasks.
Building on the success of LLMs, Vision-Language Models (VLMs) have emerged by integrating visual inputs into LLMs, enabling models to reason about images and text simultaneously~\cite{achiam2023gpt,zhu2025internvl3exploringadvancedtraining,liu2024visual,abdin2024phi}. 

VLMs typically process images through a pre-trained vision encoder, creating embeddings that are subsequently aligned with and fed into a pre-trained LLM.

\paragraph{Hallucination issues in VLMs}

Despite their significant progress, VLMs often generate text that is semantically coherent but conflicts with the content of the input image, a problem generally referred to as ``hallucinations''~\cite{rohrbach2018object, liu2024survey}. 
Causes of VLMs' hallucinations are diverse, including biases in fine-tuning data~\cite{li2023evaluating}, image encoder limitations~\cite{ganz2024question, tong2024eyes}, and language biases in the LLM decoder~\cite{niu2021counterfactual, goyal2017making}. 
To address hallucinations, various mitigation methods and datasets~\cite{ganz2024question, tong2024eyes, li2023evaluating, gunjal2024detecting, guan2024hallusionbench, sun2023aligning} have been proposed.

Although hallucination mitigation is a popular research focus, it remains a key challenge for VLMs. In this work, we focus on a specific hallucination category, where models do not withhold answers to unanswerable visual questions.

\paragraph{Unanswerable visual question answering}

In Visual Question Answering (VQA)~\cite{antol2015vqa}, VLMs are shown an image and a related question and are expected to provide an accurate response.
While early benchmarks focused on open-ended questions~\cite{antol2015vqa,goyal2017making,krishna2017visual}, which is the most elementary and real-world version, more recent works include other VQA types, such as yes/no~\cite{fu2023mme} and multiple-choice questions~\cite{liu2024mmbench,li2023seed,yue2024mmmu}. 
Among these, multiple-choice VQA, offering a closed set of options, has become an important VQA variant and a primary testbed for assessing VLMs, with models~\cite{liu2024visual,abdin2024phi,achiam2023gpt} rigorously tested on its benchmarks.

However, most multiple-choice and open-ended VQA benchmarks contain only answerable questions. 
For the open-ended setting, this means that the question and image are compatible. For multiple-choice, it means that the question, image, and answer set are all compatible, and a correct answer exists within the set. 
This setup does not reflect real-world scenarios, where questions may be unanswerable. 

To address this gap in multiple-choice VQA, \citet{miyai2025unsolvable} introduced the Unsolvable Problem Detection (UPD) challenge, formalizing unanswerability in the multiple-choice setting.
They defined three unanswerability types (AAD, IASD, IVQD; see explanation in \cref{sec:intro}) and published the MM-UPD benchmark, containing flawed questions alongside standard answerable questions. 
VQA unanswerability here and in our work refers to clear flaws in the VQA input, and not knowledge gaps or uncertainty~\cite{chandu2025certainlyuncertain, he2024tubenchbenchmarkinglargevisionlanguage}.
Using the MM-UPD benchmark, \citet{miyai2025unsolvable} showed that VLMs often respond with an answer even when no relevant option exists, with open-source models showing particularly low performance. A similar finding was reported for finer-grained multiple-choice answerability types in~\cite{he2024tubenchbenchmarkinglargevisionlanguage}.

\begin{figure*}[!t]
    \centering
    \begin{subfigure}{0.490\linewidth}
        \includegraphics[width=\linewidth]{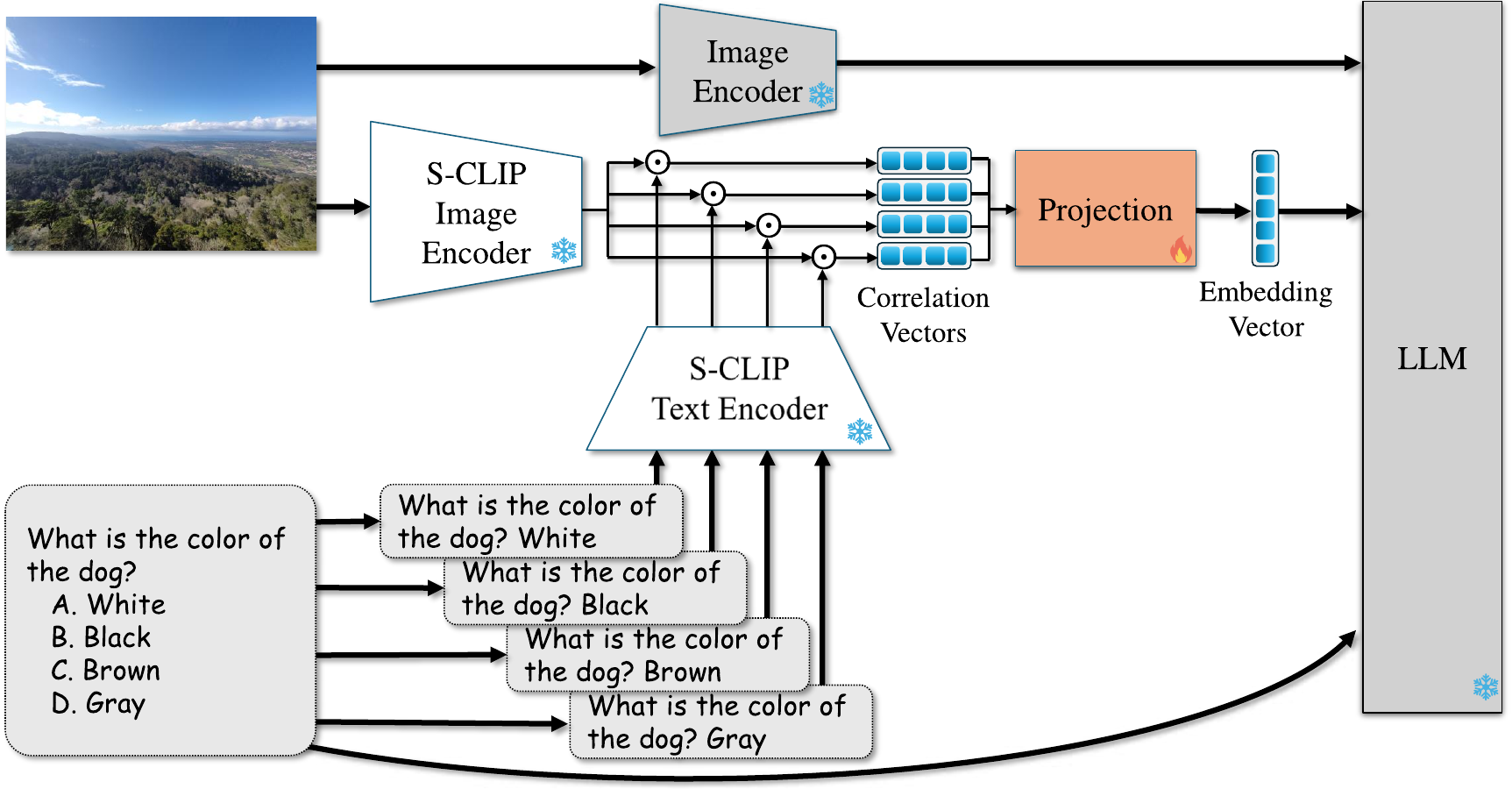}
        \caption{}
        \label{fig:clip_up_arch:multiple_choice}
    \end{subfigure}
    \begin{tikzpicture}[overlay]
        \draw[dashed, thick] (0.07, 5.0) -- (0.07, 0.2);
    \end{tikzpicture} 
    \hfill
    \begin{subfigure}{0.490\linewidth}
        \centering
        \includegraphics[width=\linewidth]{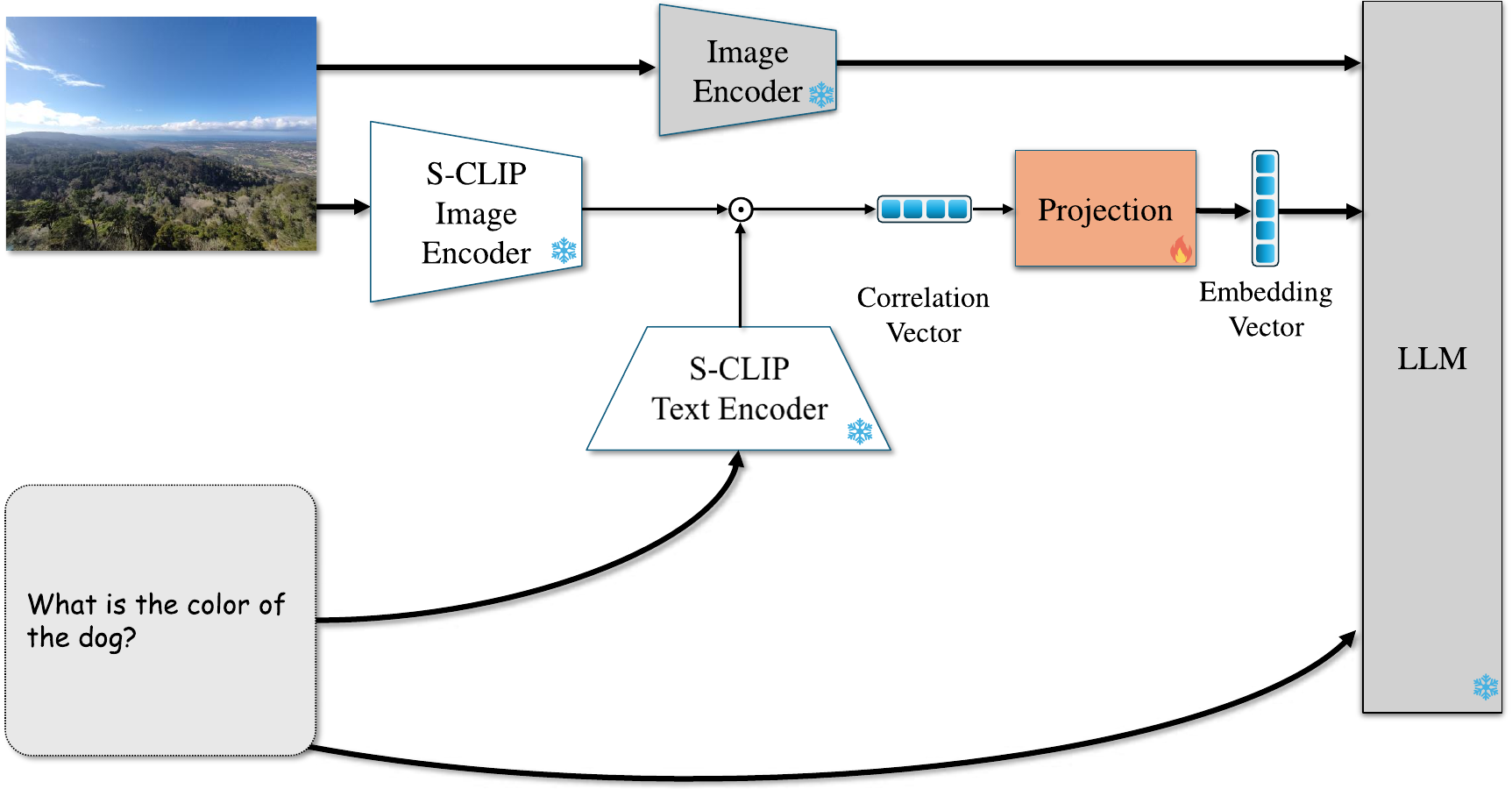}
        \caption{}
        \label{fig:clip_up_arch:open_ended}
    \end{subfigure}
  \caption{CLIP-UP embedding injection applied on common VLM architectures. 
  (a) For multiple-choice questions, given an image and a VQA prompt, the prompt is transformed into text segments merging the question with each answer option. 
  These segments and the image are encoded by Structure-CLIP (S-CLIP) to produce embeddings, from which correlation vectors are formed via element-wise multiplication. A learnable projection module maps these vectors into the VLM's intermediate feature space. The resulting new embedding vector is integrated into the LLM component of the VLM alongside the standard inputs. 
  (b) For open-ended questions, the process is similar but involves a single correlation vector computed from the image and question.}
  \label{fig:clip_up_arch}
\end{figure*}

For open-ended VQA, no prior work has formalized coarse types of unanswerability, as it appears that only one type exists, namely the incompatibility between the image and the question (parallel to IVQD in multiple-choice VQA)~\cite{guo2024unk,zhang2023toward,qian2024easyfoolmultimodalllms}. 
However, some works~\cite{gurari2018vizwiz,qian2024easyfoolmultimodalllms,chandu2025certainlyuncertain} have proposed finer-grained categories of open-ended unanswerability, such as visually deceptive images~\cite{qian2024easyfoolmultimodalllms} or ambiguity due to poor image quality~\cite{gurari2018vizwiz}.
Similar to the multiple-choice case, popular VLMs have been found to perform poorly on unanswerable open-ended questions~\cite{qian2024easyfoolmultimodalllms,chandu2025certainlyuncertain}.

In this work, we show that although improved in recent VLMs, these challenges persist and present a solution leveraging CLIP in a lightweight training framework.

\paragraph{Efficient model editing}

Model editing is a research area focused on making targeted changes to pre-trained models for specific inputs, without compromising overall performance. Efficiency is a key goal, with efforts aimed at developing methods to apply edits without high computational or extensive data requirements~\cite{yao2023editing,zhang2024comprehensive}. 

Model editing has gained attention in the context of LLMs, driven by the growing need to adapt models to issues such as updating outdated information~\cite{hartvigsen2023aging,mitchell2021fast,de2021editing}. It is also popular in text-to-image generation, where personalization may be achieved by affordable learning of concept-specific embedding vectors~\cite{ruiz2023dreambooth,gal2022image,vinker2023concept,avrahami2023break}. 

Few works have explored model editing in VLMs~\cite{alaluf2024myvlm,cheng2023edit,huang2024vlkeb}. MyVLM~\cite{alaluf2024myvlm}, for instance, learns concept-specific embedding vectors for personalized VLM outputs. Our CLIP-UP method also introduces a new embedding vector, but to enhance VLMs with the ability to withhold answers to unanswerable questions. 
Moreover, rather than learning the vector from scratch, we leverage CLIP's vision-language alignment and learn projection layers to create it.

%% file: sec/3_method.tex
\section{Method}
\label{sec:method}


This section first describes CLIP-UP embedding injection for multiple-choice questions, and then explains how it extends to open-ended questions with minimal changes. An overview of our approach applied to common VLM architectures is shown in \cref{fig:clip_up_arch:multiple_choice}.
Given an image and a multiple-choice VQA prompt, we first parse the text to form individual text segments that merge the question with each answer.
Each text segment is encoded by CLIP-based text encoder, while the image is encoded by CLIP-based image encoder. Each text segment encoding is multiplied element-wise by the image encoding to create \emph{correlation vectors}.
These vectors capture the alignment between the image and each answer option (see \cref{sec:method:signal}). Next, we concatenate the correlation vectors and pass them through a learnable projection layer, transforming them into a new embedding vector within the VLM's feature space. This embedding is then fed into the LLM component of the VLM, alongside standard inputs (see \cref{sec:method:learning}).

CLIP-UP for open-ended questions has one key difference: instead of computing multiple correlation vectors by correlating the image with answer options, it computes a single correlation vector between the image and the question (see \cref{sec:method:open_ended} and \cref{fig:clip_up_arch:open_ended}). 

Finally, we explore alternative ways of injecting the correlation vectors into the VLM. 
Specifically, we propose a novel method that injects the correlation vectors directly into LoRA layers during LoRA fine-tuning (see \cref{sec:method:injected_lora}).

\subsection{Correlation vectors generation}
\label{sec:method:signal}

The first step of CLIP-UP for multiple-choice VQA is to generate correlation vectors. Let $(T, I)$ be a multiple-choice VQA input, where $T$ represents the text (both the question and answer options) and $I$ is the image. Given such input with $n$ options, we parse $T$ into a pair $(Q, \{O_1, \dots, O_n\})$, where $Q$ is the question and $\{O_1, \dots, O_n\}$ is the set of answer options. Parsing is done with a simple rule-based algorithm; see \cref{supp:sec:parsing} in the Supplementary Material (Supp.) for details.

To contextualize each answer option $O_i$, we merge it with the question $Q$. This forms the set $\mathcal{Q}_{\text{opt}} = \{ Q + O_i \mid i = 1, \dots, n \}$, where each $s_i \in \mathcal{Q}_{\text{opt}}$ is the question followed by a single answer option. 
Note that using answers alone does not contain sufficient information, as it can lack context (\eg, ``Blue'' vs. ``What color is the dress? Blue'').

We then encode each contextualized answer option $s_i \in \mathcal{Q}_{\text{opt}}$ with Structure-CLIP~\cite{huang2023structure} text encoder to obtain a text embedding $\mathbf{v}_{s_i} = \text{SC}_{\text{text}}(s_i)$. The image is processed by Structure-CLIP image encoder, yielding the image embedding $\mathbf{v}_{I} = \text{SC}_{\text{img}}(I)$. 
Structure-CLIP~\cite{huang2023structure} is a variant of CLIP, designed to better distinguish semantically different texts with similar structures. It is well-suited for our contextualized answer options which share the same structure, namely, the same question followed by an answer option.

We generate $n$ correlation vectors $\{ \mathbf{u}_1, \dots, \mathbf{u}_n \}$ by performing element-wise multiplication between each text embedding and the image embedding: 
\begin{equation}
  \mathbf{u}_{i} = \mathbf{v}_{I} \odot \mathbf{v}_{s_i}. 
\end{equation}
Rather than relying solely on scalar CLIP-based similarity scores (dot product of embeddings), our approach constructs richer correlation vectors incorporating similarity scores (as the sum of an element-wise product is the dot product) and additional alignment information.

\begin{figure}[!t]
    \centering
    \def\signleSubFigWidth{0.48}
    
    \begin{subfigure}{\signleSubFigWidth\linewidth}
        \includegraphics[width=\linewidth]{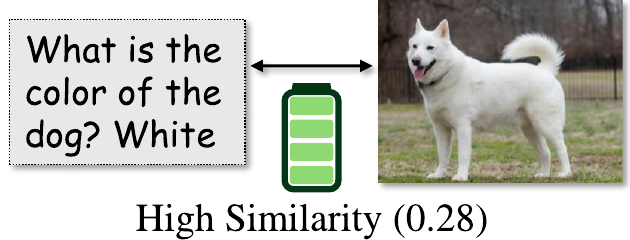}
        \caption{Standard (multiple-choice)}
        \label{fig:clip_up_signal:signal1}
    \end{subfigure}
    \hfill
    \begin{subfigure}{\signleSubFigWidth\linewidth}
        \includegraphics[width=\linewidth]{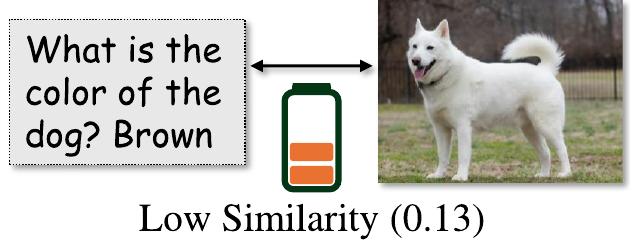}
        \caption{AAD (multiple-choice)}
        \label{fig:clip_up_signal:signal2}
    \end{subfigure}
    \vfill
    \vspace{0.15cm}
    \begin{subfigure}{\signleSubFigWidth\linewidth}
        \includegraphics[width=\linewidth]{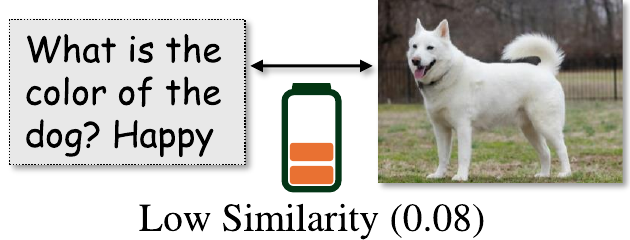}
        \caption{IASD (multiple-choice)}
        \label{fig:clip_up_signal:signal3}
    \end{subfigure}
    \hfill
    \begin{subfigure}{\signleSubFigWidth\linewidth}
        \includegraphics[width=\linewidth]{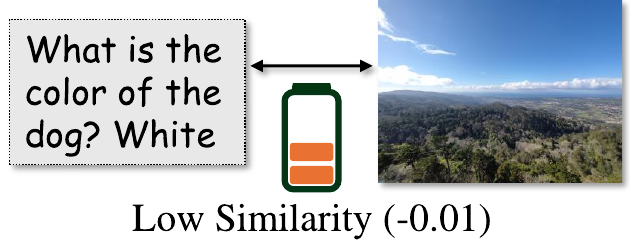}
        \caption{IVQD (multiple-choice) }
        \label{fig:clip_up_signal:signal4}
    \end{subfigure}
    \noindent\makebox[\linewidth][c]{\textbf{- - - - - - - - - - - - - - - - - - - - - - - - - - - - - - - - - - - - - - - - - -}}
    \vspace{0.15cm}
    \begin{subfigure}{\signleSubFigWidth\linewidth}
        \includegraphics[width=\linewidth]{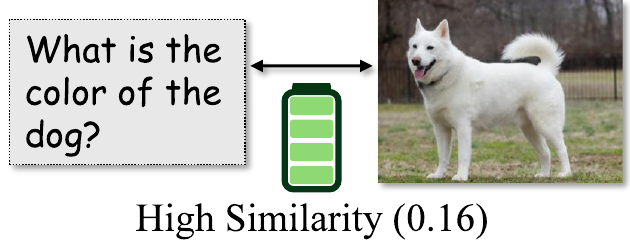}
        \caption{Standard (open-ended)}
        \label{fig:clip_up_signal:open_ended_standard5}
    \end{subfigure}
    \hfill
    \begin{subfigure}{\signleSubFigWidth\linewidth}
        \includegraphics[width=\linewidth]{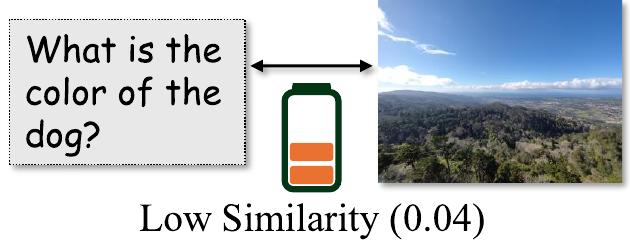}
        \caption{Unanswerable (open-ended)}
        \label{fig:clip_up_signal:open_ended_unanswerable6}
    \end{subfigure}
  \caption{Our correlation vectors capture a prior for VQA answerability. 
  In multiple-choice questions, (a) for a standard question, the correct contextualized answer option aligns well with the image, resulting in a correlation vector with high values. For unanswerable questions, no option aligns well with the image: either (b) all answer options are incorrect (AAD), (c) incorrect and irrelevant to the question (IASD), or (d) the question is incompatible with the image (IVQD). 
  Open-ended questions show a similar trend: (e) answerable questions align well with the image, in contrast to (f) unanswerable ones. Numbers represent average Structure-CLIP similarity scores measured on test data.}
  \label{fig:clip_up_signal}
\end{figure}

Extracting these correlation vectors provides a strong prior for assessing the answerability of the multiple-choice VQA input. As illustrated in Figs.~\ref{fig:clip_up_signal:signal1}--\ref{fig:clip_up_signal:signal4}, for a standard answerable question (\ie., one having a correct answer option) paired with its correct answer, the image and text align well, leading to high CLIP similarity. Consequently, for standard questions, one correlation vector, the one corresponding to the correct answer, will exhibit high values.
Conversely, for unanswerable questions, all $n$ correlation vectors are expected to exhibit low values, as no contextualized option aligns well with the image. 
Therefore the $n$ correlation vectors provide a strong signal for determining multiple-choice VQA answerability.

\subsection{Learning a new projection layer}
\label{sec:method:learning}

The correlation vectors capture essential alignment signal, but the VLM cannot directly interpret them, as they are neither optimized for its use nor aligned with its feature space dimension. To address this, we concatenate the correlation vectors and project them into a vector $\mathbf{e}$ in the VLM's feature space, using a learnable projection layer $\mathcal{P}$:
\begin{equation}
\label{eq:clip_up_projection}
    \mathbf{e} = \mathcal{P}(\mathbf{u}), \mbox{ where } \mathbf{u} = [\mathbf{u}_{1}; \ldots; \mathbf{u}_{n}].
\end{equation}

The new embedding vector $\mathbf{e}$ is subsequently fed into the LLM component of the VLM.
In particular, only the projection layer is trained, while all other components remain frozen, making CLIP-UP training simple and efficient. 
Training uses cross-entropy loss on answerable and unanswerable questions: answerable questions have the correct answer option (\eg, ``A. White'') as their ground truth text, while unanswerable questions use ``I cannot answer.''

We use a simple linear layer for the projection. Since the projection input size is fixed, the concatenated correlation vectors must have fixed size. Given that multiple-choice questions typically have up to four options, we generate four correlation vectors for each input. For inputs with fewer options (three or two), we fill the remaining slots with correlation vectors generated from element-wise multiplication of the image embedding and null text embedding. 

What makes CLIP-UP effective at enhancing UPD performance, even though VLMs already ``see'' both the image and text? We postulate that the global alignment information extracted by CLIP-UP is absent in popular VLMs~\cite{liu2023improvedllava,liu2024visual,liu2024llavanext,li2023blip,abdin2024phi,zhu2025internvl3exploringadvancedtraining}.
For example, although LLaVA uses a CLIP image encoder, it extracts patch features from the penultimate layer, assumed to be more effective for capturing image details~\cite{liu2024visual}. 
InternVL3~\cite{zhu2025internvl3exploringadvancedtraining} uses an earlier layer of its InternViT2.5 vision encoder~\cite{chen2025expandingperformanceboundariesopensource}  (layer 45 out of 48).
In contrast, CLIP-UP equips VLMs with global information by using Structure-CLIP's class embedding from the last layer, explicitly trained to capture global alignment~\cite{radford2021learning}. Together with the incorporation of Structure-CLIP's text embeddings, CLIP-UP introduces global image-text information that VLMs lack.

\begin{figure*}[!t]
    \centering
    \def\tsneSubFigWidth{0.24}

    \begin{subfigure}{\tsneSubFigWidth\linewidth}
        \includegraphics[width=\linewidth]{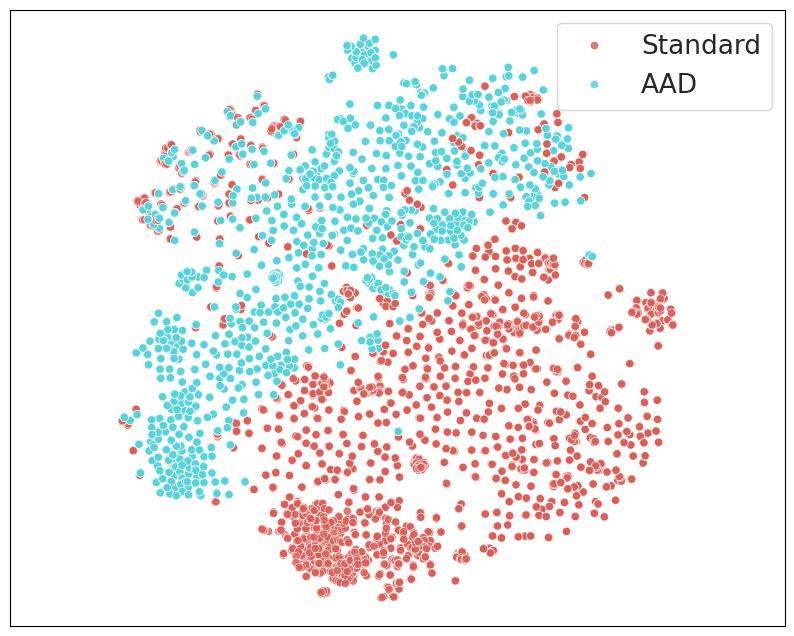}
        \caption{AAD}
        \label{fig:t_sne:AAD}
    \end{subfigure}
    \hfill
    \begin{subfigure}{\tsneSubFigWidth\linewidth}
        \includegraphics[width=\linewidth]{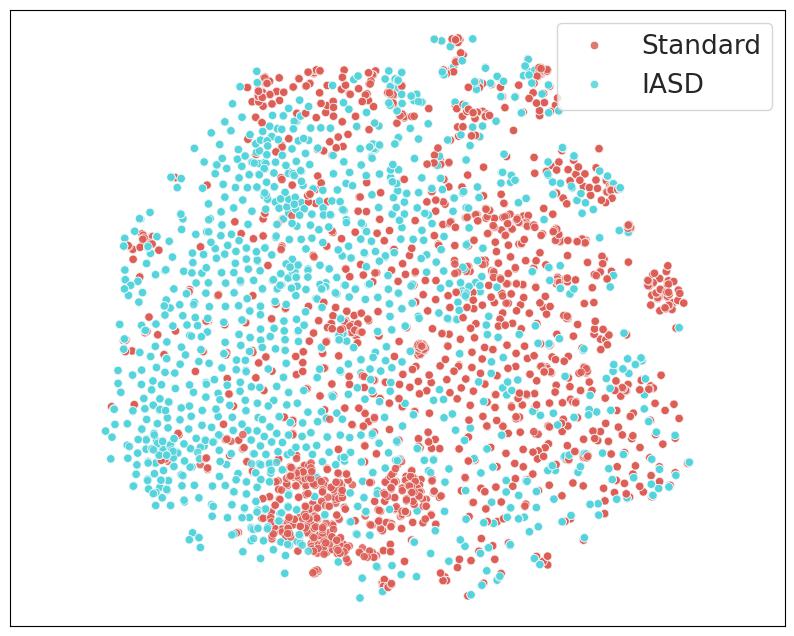}
        \caption{IASD}
        \label{fig:t_sne:IASD}
    \end{subfigure}
    \hfill
    \begin{subfigure}{\tsneSubFigWidth\linewidth}
        \includegraphics[width=\linewidth]{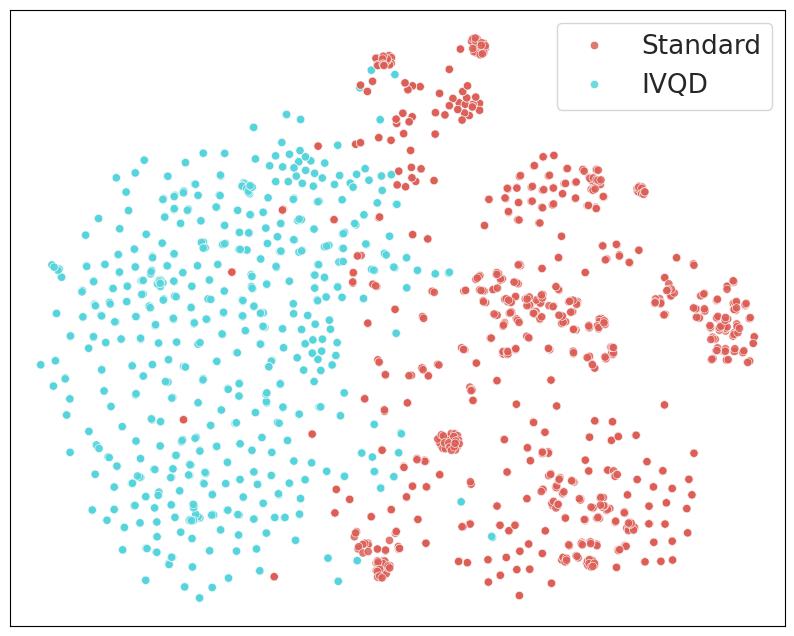}
        \caption{IVQD}
        \label{fig:t_sne:IVQD}
    \end{subfigure}
    \hfill
    \begin{subfigure}{\tsneSubFigWidth\linewidth}
        \includegraphics[width=\linewidth]{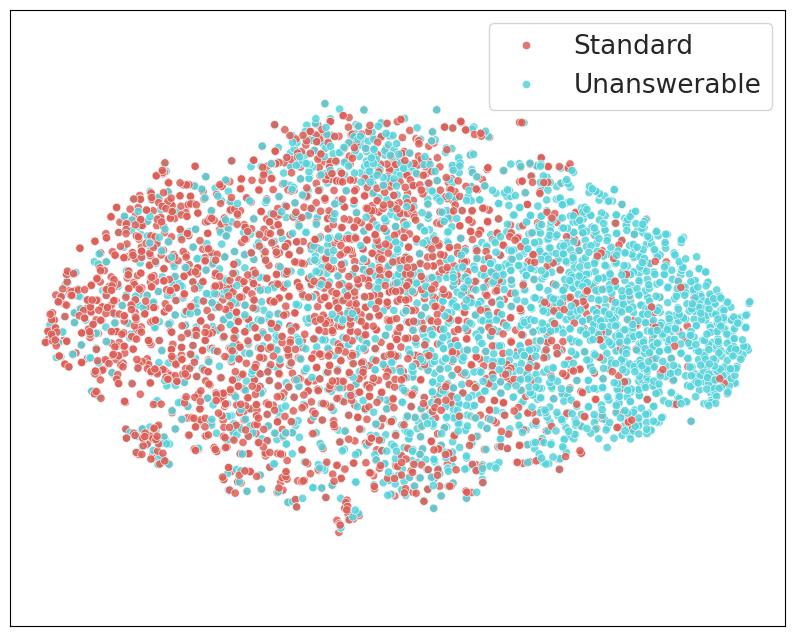}
        \caption{Open-ended}
        \label{fig:t_sne:open_ended}
    \end{subfigure}
    \caption{t-SNE plots of embedding vectors generated by CLIP-UP on LLaVA-1.5-7B, for (a-c) all samples in MM-UPD, and (d) all samples in the RGQA test data.}
    \label{fig:t_sne}
\end{figure*}

To recap, our goal is to create a new embedding vector that conveys (un)answerability information to the model. \cref{fig:t_sne} illustrates this intuition, showing that the learned projection forms distinct clusters of the embeddings of ``answerable'' and ``unanswerable'' vectors.

\subsection{Handling open-ended questions}
\label{sec:method:open_ended}

The CLIP-UP scheme proposed for multiple-choice VQA can be applied to open-ended questions. 
The key difference is that in the absence of answer options, the method computes a single correlation vector between the image and the question, expected to exhibit higher values for answerable questions (see Figs.~\ref{fig:clip_up_signal:open_ended_standard5}--\ref{fig:clip_up_signal:open_ended_unanswerable6}). 
This vector is then projected using a learnable projection layer, trained on both answerable and unanswerable open-ended questions (see~\cref{fig:clip_up_arch:open_ended}). 

Although CLIP-UP generates separate embedding vectors for multiple-choice and open-ended questions, we can flexibly activate either one or none. 
During inference, a simple classifier determines whether the input is a multiple-choice question, an open-ended question, or neither. This enables generating and using the relevant embedding vector, or omitting it entirely if the input is not a question.
Since the original VLM weights are preserved, the model performance on non-VQA tasks remains unaffected.

\subsection{Injecting correlation vectors into LoRA}
\label{sec:method:injected_lora}

The core idea of CLIP-UP is to inject an answerability prior into the VLM's decision process. So far, this has been achieved by injecting a new embedding vector into the VLM directly. However, other injection approaches are possible.
In particular, we propose a general method to inject information directly into LoRA layers during LoRA fine-tuning~\cite{hu2021lora}, referred to as \emph{Injected LoRA (InjLoRA)}, and use it to inject our correlation vectors.
InjLoRA provides a more direct and expressive injection mechanism to modulate LoRA weights than prior work. For example, \citet{stracke2024ctrloralter} inject signals via shift and scale operations.

In standard LoRA fine-tuning, pretrained weights $W_0 \in \mathbb{R}^{d \times k}$ are modified to $W_0 + \Delta W$, where $\Delta W = BA$, with $B \in \mathbb{R}^{d \times r}$, $A \in \mathbb{R}^{r \times k}$, and $r$ denoting the LoRA rank.
In our InjLoRA, we learn a projection layer $\mathcal{P'}$ for each LoRA layer, such that $\Delta W$ becomes:
\begin{equation}
\label{eq:injected_lora}
    \Delta W = B\left(\mathcal{P'}(\mathbf{u}) + C\right)A,
\end{equation}
where $\mathbf{u}$ is the concatenated correlation vectors (from \cref{eq:clip_up_projection}), and $\mathcal{P'}(\mathbf{u}) \in \mathbb{R}^{r \times r}$. The matrix $C \in \mathbb{R}^{r \times r}$ is learnable and acts as a residual, allowing LoRA layers to ignore the injected signal if desired.

We learn a different projection $\mathcal{P'}$ for each LoRA layer, allowing each layer to attend to different aspects of the correlation signal according to its role in the model.

%% file: sec/4_experiments.tex
\section{Experiments}
\label{sec:experiments}

\subsection{Datasets}
\label{sec:experiments:dataset}

\paragraph{Multiple-choice VQA}
We created a multiple-choice VQA training dataset, containing answerable and unanswerable questions across AAD, IASD, and IVQD. The dataset is organized into question pairs, each consists of a standard question and its corresponding unanswerable variant. For example, an AAD unanswerable question is generated by removing the correct option from a standard question. 
The dataset includes 293, 189, and 307 question pairs for AAD, IASD, and IVQD, respectively. For each category, 30 pairs are allocated for validation, and the rest used for training. The dataset will be released. See more details in~\cref{supp:sec:training_datasets:multiple_choice}.

\paragraph{Open-ended VQA}
For open-ended VQA training, we use 700 answerable-unanswerable question pairs sampled from the TDIUC training dataset~\cite{kafle2017analysis}, with 50 pairs assigned for validation and the rest for training.

\subsection{Training and models}
\label{sec:experiments:setup}

We evaluate two CLIP-UP methods: 
(1) CLIP-UP with embedding injection only (\emph{CLIP-UP-Emb}), and (2) CLIP-UP with both embedding injection and InjLoRA fine-tuning (\emph{CLIP-UP-EmbLoRA}).
To avoid confusion, we use the term CLIP-UP hereafter to refer to the general framework, while CLIP-UP-Emb and CLIP-UP-EmbLoRA denote the specific implementations.
Note that although we evaluate InjLoRA alongside embedding injection (in CLIP-UP-EmbLoRA), injection into LoRA layers via InjLoRA can also be used independently.

Experiments are conducted on six VLMs with diverse architectures, scales, and initial UPD ability: LLaVA-1.5-7B~\cite{liu2023improvedllava}, LLaVA-NeXT-13B~\cite{liu2024llavanext}, Phi-3.5-Vision~\cite{abdin2024phi}, Ovis2-16B~\cite{lu2024ovisstructuralembeddingalignment}, InternVL3-1B, and InternVL3-8B~\cite{zhu2025internvl3exploringadvancedtraining}.

We train CLIP-UP separately on multiple-choice and open-ended data. 
In both cases, training is conducted for 3 epochs with a cross-entropy loss. 
We train with batches containing pairs of answerable and unanswerable questions (\eg, batch size of 8 with 4 pairs), which we found to improve training stability. See~\cref{supp:sec:implementation} for additional details.

\begin{table*}[!t]
    \centering
    \setlength{\tabcolsep}{3.8pt}
    \begin{tabular}{c|ccc|ccc|ccc|ccc}
        \toprule
        \multirow{2}{*}{\textbf{Method}} &
        \multicolumn{3}{c|}{\textbf{LLaVA-1.5-7B}} &
        \multicolumn{3}{c|}{\textbf{Phi-3.5-Vision}} & 
        \multicolumn{3}{c|}{\textbf{InternVL3-8B}} &
        \multicolumn{3}{c}{\textbf{Ovis2-16B}} \\
        & Stand. & UPD & Dual &
        Stand. & UPD & Dual & 
        Stand. & UPD & Dual & 
        Stand. & UPD & Dual \\
        \midrule
        \multicolumn{1}{l|}{Original Model} & 
        67.73 & 0.11 & 0.11 &
        78.58 & 0.41 & 0.41 &
        88.95 & 0.00 & 0.00 &
        88.38 & 9.97 & 9.65 \\
        
        \multicolumn{1}{l|}{Base Setting} & 
        66.33 & 7.23 & 4.87 &
        77.12 & 0.92 & 0.89 & 
        86.32 & 44.46 & 39.45 &
        87.03 & 46.46 & 42.44 \\
        
        \multicolumn{1}{l|}{Additional-Option} & 
        65.96 & 51.25 & 38.11 &
        77.52 & 39.77 & 32.73 & 
        88.81 & 66.50 & 60.06 &
        87.52 & 70.89 & 63.57 \\
        
        \multicolumn{1}{l|}{Additional-Instruction} & 
        65.88 & 43.61 & 31.39 &
        75.80 & 50.32 & 38.99 & 
        86.44 & 80.15 & 70.44 &
        87.01 & 80.15 & 71.69 \\
        
        \multicolumn{1}{l|}{LoRA Fine-Tuning} & 
        62.60 & 76.70 & 50.25 &
        60.02 & 86.99 & 53.69 &
        86.28 & 82.10 & 71.59 &
        85.88 & 87.62 & 76.65 \\
        
        \multicolumn{1}{l|}{CLIP-UP-Emb (ours)} & 
        59.72 & 80.30 & 51.22 &
        60.96 & 87.82& 54.07 &
        84.09 & 90.98 & 77.62 &
        83.50 & 91.63 & 77.92 \\
        
        \multicolumn{1}{l|}{CLIP-UP-EmbLoRA (ours)} & 
        60.72 & 83.34 & \textbf{53.07} & 
        70.17 & 86.62 & \textbf{62.27} &
        85.83 & 90.88 & \textbf{79.22} &
        87.48 & 90.20 & \textbf{80.59} \\
        
        \bottomrule
    \end{tabular}
    \caption{
    Results (\%) on MM-UPD~\cite{miyai2025unsolvable} multiple-choice VQA. Metrics include circular standard, UPD, and dual accuracies.
    }
    \label{table:multiple_choice_results}
\end{table*}

\begin{table*}[!t]
    \centering
    \setlength{\tabcolsep}{3.8pt}
    \begin{tabular}{c|ccc|ccc|ccc|ccc}
        \toprule
        \multirow{2}{*}{\textbf{Method}} &
        \multicolumn{3}{c|}{\textbf{LLaVA-1.5-7B}} &
        \multicolumn{3}{c|}{\textbf{Phi-3.5-Vision}} & 
        \multicolumn{3}{c|}{\textbf{InternVL3-8B}} &
        \multicolumn{3}{c}{\textbf{Ovis2-16B}} \\
        & Stand. & UPD & Dual &
        Stand. & UPD & Dual & 
        Stand. & UPD & Dual & 
        Stand. & UPD & Dual \\
        \midrule
        \multicolumn{1}{l|}{Original Model} & 
        68.40 & 11.08 & 7.22 & 
        69.76 & 9.20 & 6.72 &
        69.84 & 14.80 & 10.88 &
        69.56 & 12.18 & 8.68 \\
        
        \multicolumn{1}{l|}{Prompt Engineering} &
        67.24 & 14.52 & 9.56 &
        69.30 & 25.44 & 18.06 &
        67.26 & 55.22 & 35.90 & 
        29.44 & 94.40 & 26.58 \\
        
        \multicolumn{1}{l|}{LoRA Fine-Tuning} &
        57.44 & 63.14 & 31.78 &
        68.66 & 47.60 & 32.20 &
        67.14 & 61.74 & 39.84 &
        64.62 & 68.18 & 41.98 \\
        
        \multicolumn{1}{l|}{CLIP-UP-Emb (ours)} &
        39.74 & 77.14 & 28.42 &
        60.52 & 60.00 & 32.96 &
        67.04 & 57.58 & 38.42 &
        61.46 & 70.46 & 41.56 \\
        
        \multicolumn{1}{l|}{CLIP-UP-EmbLoRA (ours)} &
        57.64 & 60.64 & \textbf{32.36} &
        67.74 & 61.26 & \textbf{40.28} &
        66.06 & 67.04 & \textbf{42.88} &
        62.58 & 74.10 & \textbf{44.52} \\

        \bottomrule
    \end{tabular}
    \caption{Results (\%) on RGQA~\cite{zhang2023toward} open-ended VQA. Metrics include standard, UPD, and dual accuracies.
    }
    \label{table:rgqa_open_ended_results_main}
\end{table*}

\subsection{Method comparisons}
\label{sec:experiments:comparisons}

\paragraph{Multiple-choice VQA}
We first compare CLIP-UP to the original VLMs, where multiple-choice VQA is evaluated with the instruction ``Answer directly with the letter of the correct option from the given choices'' added to the prompt~\cite{miyai2025unsolvable}.
We also compare to prior UPD solutions, including prompt engineering settings from~\cite{miyai2025unsolvable}: Base Prompt, with no added instructions, and Additional-Option and Additional-Instruction which encourage VLMs to withhold answers to unanswerable questions. 
For example, Additional-Option adds an option indicating that no answer is correct (see~\cref{supp:subsec:additional_baselines}).
Finally, we compare CLIP-UP to LoRA fine-tuning~\cite{hu2021lora} following~\cite{miyai2025unsolvable}. 
To ensure strict evaluation, we use each VLM's recommended LoRA settings, which are more expressive (up to rank 128) than those used in our 
multiple-choice experiments with InjLoRA (rank 8).

\paragraph{Open-ended VQA}

For open-ended questions, we first compare to the original models, evaluated using the instruction ``Answer the question using a single word or phrase'' as in~\cite{liu2023improvedllava}. 
We also test prompt engineering by adding ``When the provided information is insufficient, respond with `I cannot answer'.'' to the original prompt. 
Finally, we compare CLIP-UP methods to LoRA fine-tuning, as it is the most effective baseline for multiple-choice questions.

\subsection{Benchmarks and evaluation metrics}
\label{sec:experiments:benchmarks_and_metrics}

\paragraph{Multiple-choice experiments}
\label{sec:experiments:benchmarks_and_metrics:upd}

We evaluate CLIP-UP on the comprehensive MM-UPD benchmark~\cite{miyai2025unsolvable}, which consists of three sub-benchmarks for AAD, IASD, and IVQD. 
Each sub-benchmark contains pairs of multiple-choice VQA questions, an answerable one and an unanswerable one. To account for options' order variation, the questions are repeated $n$ times ($n$ is the number of options), with a different circular shift applied to the options each time. 

Evaluation uses three metrics: circular standard accuracy, circular UPD accuracy, and circular dual accuracy.
Circular accuracy~\cite{liu2024mmbench} evaluates accuracy across all circular shifts of options by counting success only if all shifts are answered correctly.
In particular, circular standard and UPD accuracies measure the circular accuracy for answerable and unanswerable questions, respectively.

Circular dual accuracy~\cite{miyai2025unsolvable} requires correctly answering all circular variants of both standard questions and their unanswerable pairs. 
We use it as our main metric, as high dual accuracy indicates consistent discernment of answerability, in addition to correctly answering answerable questions.
Note that, in contrast, maximizing either standard or UPD accuracy individually is trivial (by using the VLM as-is or by always refraining from answering, respectively).

\paragraph{Open-ended experiments}

We evaluate open-ended VQA on 5{,}000 examples from each of the RGQA benchmark~\cite{zhang2023toward} and TDIUC test set~\cite{kafle2017analysis}, each consisting of answerable-unanswerable question pairs.
We report standard, UPD, and dual accuracies, all measured in regular form.
Evaluation follows the LAVE metric~\cite{manas2024improving}, which leverages LLMs for open-ended VQA scoring.

\subsection{Results}
\label{sec:experiments:results}

\paragraph{Multiple-choice results}
\label{sec:experiments:results:upd}

\cref{table:multiple_choice_results} presents the MM-UPD results of CLIP-UP methods applied to LLaVA-1.5-7B, Phi-3.5-Vision, InternVL3-8B, and Ovis2-16B averaged over AAD, IASD, and IVQD.
\cref{table:full_mm_upd_results1_supp,table:full_mm_upd_results2_supp} (in Supp.) show the full results, including those for other VLMs.

Although recent VLMs have improved (\eg Ovis2-16B), large gaps remain between standard and dual accuracies.
CLIP-UP-EmbLoRA addresses this, achieving the best dual accuracy with gains up to $8.58\%$ over LoRA fine-tuning. 
For example, it improves performance by $7.63\%$ on the strong InternVL3-8B, 
demonstrating that it can effectively address the UPD problem.
CLIP-UP-Emb also achieves strong performance, surpassing all baselines in most 
cases, further showing the effectiveness of the prior injection.

Although CLIP-UP methods improve UPD performance, the lightweight training is not intended to boost standard accuracy. The standard accuracy of the original model therefore serves as an upper bound for dual accuracy. 
Thus, CLIP-UP is closer to its upper limit than it may seem. For example, on Ovis2-16B, CLIP-UP-EmbLoRA achieves $80.59\%$ dual accuracy, where the upper bound is $88.38\%$.

However, this upper bound may not reflect the true knowledge of the model, as the original model may rely on answer elimination shortcut strategy.
Since CLIP-UP discourages this strategy, the standard accuracy with CLIP-UP may better reflect the model’s actual understanding. 

Improving dual accuracy comes at the cost of reduced standard accuracy, reflecting a trade-off. 
In some cases, standard accuracy drop is minimal (\eg, Ovis2-16B), but in others, depending on the application, drops may be too steep. 
To address this, in~\cref{supp:sec:inference_time_control} we introduce a simple inference-time method for controlling this trade-off.

\begin{table}[!t]
    \centering
    \setlength{\tabcolsep}{3.7pt}
    \begin{tabular}{c|ccc}
        \toprule
        \multirow{1}{*}{\textbf{Method}} & 
        Stand. & UPD & Dual \\
        \midrule
        
        \multicolumn{1}{l|}{CLIP-UP-Emb w\slash{} const. signal} &
        56.82 & 75.17 & 45.27 \\
        
        \multicolumn{1}{l|}{CLIP-UP-Emb w\slash{} similarities} &
        60.39 & 73.84 & 47.90 \\
        
        \multicolumn{1}{l|}{CLIP-UP-Emb w\slash{} CLIP ViT-L/14} &
        47.94 & 80.97 & 39.00 \\
        
        \multicolumn{1}{l|}{CLIP-UP-Emb (ours)} & 
        59.72 & 80.30 & \textbf{51.22} \\

        \hline
        
        \multicolumn{1}{l|}{CLIP-UP-Emb + standard LoRA} &
        59.84 & 83.35 & 52.42 \\
        
        \multicolumn{1}{l|}{CLIP-UP-EmbLoRA (ours)} &
        60.72 & 83.34 & \textbf{53.07} \\
        
        \bottomrule
    \end{tabular}
    \caption{Ablation results (\%) on LLaVA-1.5-7B.}
    \label{table:ablation}
\end{table}

\paragraph{Open-ended results} 

For open-ended questions, ~\cref{table:rgqa_open_ended_results_main,table:rgqa_open_ended_results_sup}
 present the RGQA results, and \cref{table:tdiuc_open_ended_results} shows results on TDIUC.
On RGQA, CLIP-UP-EmbLoRA surpasses LoRA in almost all cases with gains up to $8.08\%$. TDIUC shows similar trends, with gains up to $11.32\%$.

\paragraph{Classification of question type}
\label{sec:experiments:classiciation_of_question_type}

We address the real-world case where inputs may be multiple-choice questions, open-ended questions, or neither.
CLIP-UP requires knowing the input type, to determine which embedding to generate.
We find that a simple rule-based classifier (described in~\cref{supp:sec:parsing}) is highly effective: 
it correctly identifies all tested multiple-choice and open-ended inputs, and classifies others (\eg, requests to caption an image) as non-questions. 
While unnecessary for our data, classification can be done in other ways, such as using the VLM's LLM component.

\subsection{Ablation study and limitations}
\label{sec:experiments:ablations}

\cref{fig:t_sne} visualizes the projections learned by CLIP-UP-Emb with t-SNE plots of embedding vectors generated from all samples in the three MM-UPD sub-benchmarks and RGQA test data.
As CLIP-UP is intended to discern answerable questions from unanswerable ones, we expect t-SNE to reveal distinct clusters for these two groups. 
Indeed, clustering appears for all data types: AAD and IVQD clustering is clearest, while IASD clustering is less distinct, likely because the correlation vectors are less suited to capture its textual inconsistency (\ie, question-option mismatches). 
Open-ended clustering is also less clear, presumably due to the task difficultly (see discussion below).

We perform ablation studies to assess the impact of CLIP-UP's components, with results for CLIP-UP-Emb on MM-UPD shown in \cref{table:ablation}.
We examine the impact of using Structure-CLIP by replacing the correlation vectors with a constant signal, effectively learning an embedding vector from scratch (line 1 in \cref{table:ablation}).
This approach is inferior to using Structure-CLIP, confirming that it is a key factor in CLIP-UP and that the gains are not solely due to learning a new embedding vector.
Interestingly, the results are not as low as may be expected, presumably because the learned vector encourages the model to extract alignment information already present internally, while the correlation vectors provide additional cues not otherwise accessible.

We examine a CLIP-UP variant where the projection is based on four scalar Structure-CLIP similarities rather than on four correlation vectors (line 2).
While this approach performs well, it yields weaker results than CLIP-UP, underscoring the value of the richer information within the correlation vectors.
We additionally test CLIP-UP using CLIP ViT-L/14 to generate the correlation vectors instead of Structure-CLIP (line 3), 
and observe markedly lower results compared to our full CLIP-UP-Emb, underscoring Structure-CLIP's importance.
Interestingly, the performance is higher with constant signal (line 1) than with the CLIP ViT-L/14. 
We hypothesize that the optimization process requires the signal to be highly correlated with the task; otherwise, a conservative solution using the same signal for all inputs may be more effective.

We also test whether CLIP-UP-EmbLoRA benefits from the injection into LoRA layers, or merely from adding LoRA fine-tuning. 
We observe that it outperforms CLIP-UP-Emb with standard LoRA (lines 5 and 6), highlighting the benefit provided by injecting priors into LoRA.

Finally, while CLIP-UP offers a comprehensive solution for multiple-choice, it may be limited in addressing the diversity of open-ended questions. 
For instance, it is not expected to help with highly general open-ended prompts (\eg, ``What is in this image?'').
Another potential limitation stems from the reliance on Structure-CLIP signals, which may be unsuitable in some cases.
Please see~\cref{supp:sec:limitations} for further discussion of limitations.

%% file: sec/5_conclusions.tex
\section{Conclusion}
\label{sec:discussion}

This paper introduces CLIP-UP, a lightweight approach for enhancing pre-trained VLMs' ability to withhold responses to unanswerable open-ended and multiple-choice VQA questions. 
CLIP-UP leverages CLIP-based measures to learn a few linear projections to achieve this, without altering the original VLM weights. 
Evaluated across several models, CLIP-UP achieves consistent gains over other methods while preserving performance on non-VQA tasks.

Beyond improving models' robustness on unanswerable questions, CLIP-UP also contributes to more trustworthy evaluation of VLMs, as it discourages the strategy of eliminating unlikely options. Thus, it may better uncover VLMs' true knowledge. 
While this contribution is demonstrated in the vision-language setting, the approach could be similarly applied to other modalities, for example, by 
leveraging CLAP~\cite{wu2023large} instead of CLIP in audio-language models.

%% file: sec/acknowledgements.tex
\section*{Acknowledgments}
\label{sec:acknowledgments}

This research was partially supported by Joint NSFC-ISF Research Grant no.~\mbox{3077/23} and Israel Science Foundation Grant no.~\mbox{1427/25}.


%% file: sec/X_suppl.tex
\maketitlesupplementary
\appendix

\section{Implementation details}
\label{supp:sec:implementation}

\subsection{CLIP-UP-Emb}
\label{supp:subsec:implementation_clip_up_only}

We begin by describing the setup used to train CLIP-UP-Emb. Unless otherwise specified, the training settings are identical across both multiple-choice and open-ended tasks, and across all evaluated VLMs.

We use the AdamW optimizer~\cite{loshchilov2018decoupled} with a weight decay of 0.0001, 
a cosine learning rate schedule~\cite{loshchilov2017sgdr}, and a total of 3 training epochs. The learning rate starts at 0.0625 and decays to 0 over the course of training.
We use a batch size of 8 for LLaVA-1.5-7B, Phi-3.5-Vision and InternVL3-1B, and a batch size of 4 for LLaVA-NeXT-13B, Ovis2-16B, and InternVL3-8B.
Gradient checkpointing is applied in all settings to reduce GPU memory usage. 
For Phi-3.5-Vision, we additionally apply gradient clipping to a maximum norm of 0.5.

As we observed that training produces separability between embedding vectors from answerable and unanswerable inputs (see~\cref{fig:t_sne}), we begin multiple-choice training with one warm-up epoch using supervised contrastive loss~\cite{khosla2020supervised}, aiming to separate ``answerable'' and ``unanswerable'' projected embeddings. 
This stage uses a batch size of 128, a temperature of 0.07, and a constant learning rate of 0.0005. 
Open-ended training skips the warm-up phase.

LLaVA models use float16 precision, while other models use bfloat16 precision.
Phi-3.5-Vision, InternVL3-1B/8B and Ovis2-16B were set to process up to 4, 6, and 1 image crops, respectively.

We generate the correlation vectors using Structure-CLIP~\cite{huang2023structure}. Its embedding dimension is 768, resulting in correlation vectors with a total dimension of 3072 for 
multiple-choice VQA (with four concatenated correlation vectors) and 768 for open-ended VQA (with a single correlation vector).
The learned linear projection layer includes a bias term and operates in bfloat16 precision.

For multiple-choice VQA, training and inference are conducted using the base prompt setting, where the model receives the VQA inputs without additional instructions. For open-ended questions, the instruction ``Answer the question using a single word or phrase'' is included. All experiments are performed using greedy decoding.

Training was done on two NVIDIA GeForce RTX 3090 GPUs for LLaVA-1.5-7B, Phi-3.5-Vision, Ovis2-16B and InternVL3 models, and on a single NVIDIA RTX A6000 GPU for LLaVA-NeXT-13B.

\subsection{CLIP-UP-EmbLoRA}
\label{supp:subsec:implementation_injlora}

We describe the CLIP-UP-EmbLoRA setting. CLIP-UP-EmbLoRA combines both embedding injection and InjLoRA fine-tuning.
For training the embedding injection, we use the same CLIP-UP-Emb configuration described above.

For InjLoRA, projections ($\mathcal{P'}$ in~\cref{eq:injected_lora}) are trained using the same settings, and LoRA fine-tuning is applied to all linear layers in the LLM component of the VLM.
For all the LoRA layers and the learnable residual matrices ($C$ in~\cref{eq:injected_lora}), we use a cosine learning rate schedule that decays to 0, with a linear warmup over the first 3\% of training steps. 
Learning rate starts at $1 \times 10^{-5}$ for LLaVA models and Phi-3.5-Vision, $4 \times 10^{-5}$ for InternVL3 models, and $1 \times 10^{-4}$ for Ovis2-16B.
For multiple-choice VQA, we use a LoRA rank of 8 and a LoRA alpha of 16, while for open-ended VQA, we use a rank of 32 and an alpha of 64. 

We note that since InjLoRA introduces signal-dependent modifications to the LoRA layers, the LoRA weights cannot be merged into the base model at inference time.

\subsection{Structure-CLIP}
We generate the correlation vectors using Structure-CLIP~\cite{huang2023structure}.
As Structure-CLIP’s weights are not published, we fine-tune CLIP ViT-L/14@336~\cite{radford2021learning} to replicate it.

Fine-tuning is performed on CLIP ViT-L/14@336px~\cite{radford2021learning} for one epoch on a single NVIDIA A100 GPU, over the MS COCO dataset~\cite{lin2014microsoft} with augmentations by~\cite{huang2023structure}. 
To reduce memory usage, we freeze the first 9 transformer blocks of the image encoder and the first 21 transformer blocks of the text encoder.
The Knowledge-Enhanced Encoder (KEE) component is fine-tuned following the procedure in~\cite{huang2023structure}.
We use a learning rate of $3 \times 10^{-6}$, a batch size of 16, a weight decay of 0.1, and a KEE Knowledge weight of 0.2.
In inference, we use the fine-tuned image and text encoders of Structure-CLIP without the additional KEE.

\subsection{LoRA fine-tuning baseline}

In the main paper, we compare CLIP-UP methods to LoRA fine-tuning~\cite{hu2021lora}.
For multiple-choice UPD on LLaVA models, we followed the setup proposed in~\cite{miyai2025unsolvable}: 
we used their published LLaVA-NeXT-13B weights, and for LLaVA-1.5-7B, which was not evaluated in their work, we reproduced the fine-tuning process using their LLaVA-NeXT-13B settings and training data.

Phi-3.5-Vision, InternVL3 models and Ovis2-16B were neither fine-tuned in~\cite{miyai2025unsolvable},
so we followed their recommended LoRA training recipe:
for Phi-3.5-Vision, a learning rate of $2 \times 10^{-4}$, batch size of 64, LoRA rank of 32, and LoRA alpha of 16; 
for InternVL3 models, a learning rate of $4 \times 10^{-5}$, batch size of 64, LoRA rank of 16, and LoRA alpha of 32;
for Ovis2-16B, a learning rate of $1 \times 10^{-4}$, batch size of 4, LoRA rank of 32, and LoRA alpha of 64.

For open-ended VQA, we applied the multiple-choice recipes with our own training data.

\begin{figure*}[!t]
    \centering
    \begin{subfigure}{0.95\linewidth}
        \centering
        \includegraphics[width=\linewidth]{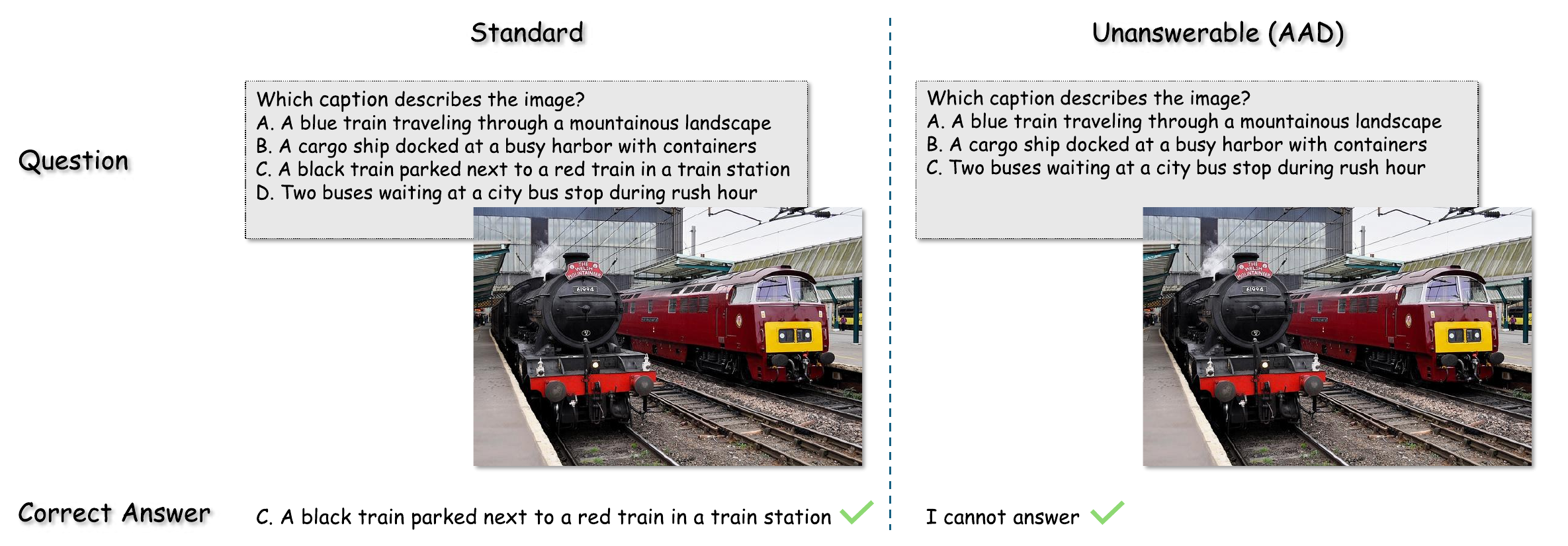}
        \caption{}
        \label{fig:supp:dataset_examples:AAD}
    \end{subfigure}
    \vfill
    \vspace{0.35cm}
    \begin{subfigure}{0.95\linewidth}
        \centering
        \includegraphics[width=\linewidth]{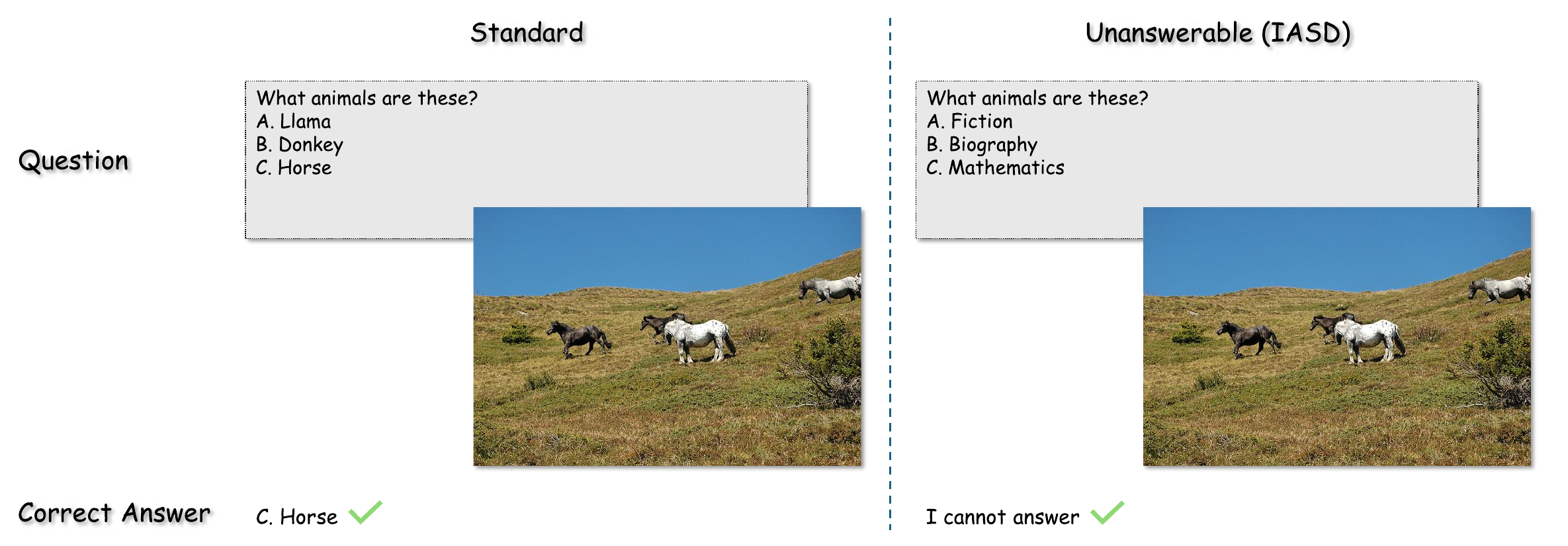}
        \caption{}
        \label{fig:supp:dataset_examples:IASD}
    \end{subfigure}
    \vfill
    \vspace{0.35cm}
    \begin{subfigure}{0.95\linewidth}
        \centering
        \includegraphics[width=\linewidth]{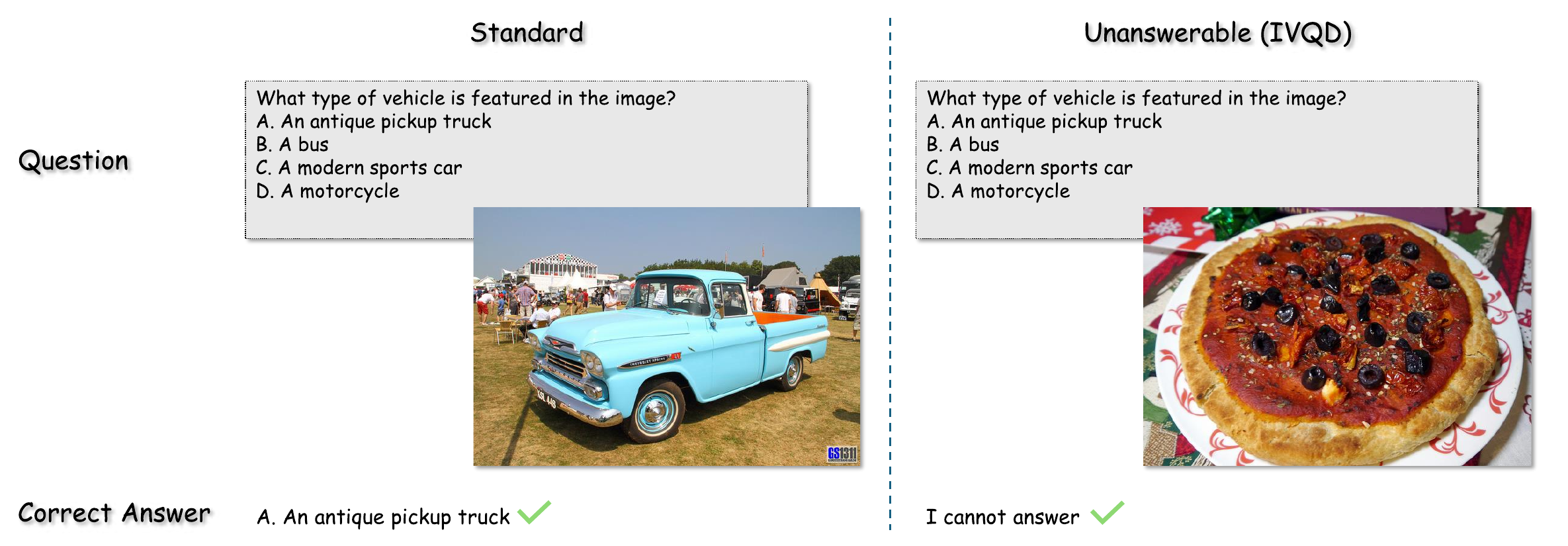}
        \caption{}
        \label{fig:supp:dataset_examples:IVQD}
    \end{subfigure}
    \caption{Pairs of standard and unanswerable multiple-choice VQA questions from our multiple-choice dataset for (a) AAD, (b) IASD, and (c) IVQD.}
    \label{fig:supp:dataset_examples}
\end{figure*}

\begin{figure*}[t]
    \centering
    \begin{subfigure}{\linewidth}
        \centering
        \fcolorbox{black}{gray!10}
        {
            \begin{minipage}{0.9\linewidth}
                You are an assistant with the task of creating multiple-choice questions about images. You will be given an image, and its correct caption. The correct caption is the correct answer to the question ``Which one is the correct caption of this image?''.\\
                Your job is to create 3 distractors that are incorrect captions for the image. Note that the distractors must be incorrect. This means that if we will take off the correct option, there will be no correct distractor that might describe the image.\\
                The output should be in the form of a python dictionary, with 6 entries: \doublestraightquotes{question} containing the question, \doublestraightquotes{image\_id} containing an image id as integer (that will be given as input), \doublestraightquotes{A} containing the correct caption, and \doublestraightquotes{B}, \doublestraightquotes{C}, \doublestraightquotes{D} containing (each) the 3 distractors.\\
                Here is an output for example:
                \{\doublestraightquotes{question}: \doublestraightquotes{Which one is the correct caption of this image?}, 
                \doublestraightquotes{image\_id}: 57703, 
                \doublestraightquotes{A}: \doublestraightquotes{A man and two women walking their dogs and hiking in the woods.}, 
                \doublestraightquotes{B}: \doublestraightquotes{A group of people camping near a lake with their pets.}, 
                \doublestraightquotes{C}: \doublestraightquotes{Three hikers climbing a mountain trail with no animals in sight.}, 
                \doublestraightquotes{D}: \doublestraightquotes{Two women and a child having a picnic in a grassy field.}\}
            \end{minipage}
        }
        \caption{}
        \label{fig:supp:gpt_instructions:aad}
    \end{subfigure}
    \vfill
    \vspace{0.3cm}
    \begin{subfigure}{\linewidth}
        \centering
        \fcolorbox{black}{gray!10}
        {
            \begin{minipage}{0.9\linewidth}
            You are an assistant with the task of creating a ``specific'' question about an image. You will be given a caption of an image (without the image itself), and you should phrase a question that can be answered using the information in this caption. The question must be phrased so it delivers some information about the image, thus it will not be relevant for any image. In addition, the information in the caption must be necessary to answer the question. You may deliver only some information about the caption, and not all of it, use your judgment. Please try to output long answers when possible.\\
            The output should be in the form of a python dictionary, with 3 entries: \doublestraightquotes{image\_id} containing an image id as integer (that will be given as input), \doublestraightquotes{question} containing the question, and \doublestraightquotes{answer} containing the answer.\\
            For you to understand, here are some examples. Each example contains input and output, an additional undesired output with an explanation:\\
            Example 1:\\
            Input: \{\doublestraightquotes{image\_id}: 32677, \doublestraightquotes{caption}: \doublestraightquotes{A dog and a cat sleeping next to each other.}\}\\
            Output: \{\doublestraightquotes{image\_id}: 32677, \doublestraightquotes{question}: \doublestraightquotes{What animals are sleeping in the image?}, \doublestraightquotes{answer}: \doublestraightquotes{A dog and a cat.}\}\\
            Undesired output: \{\doublestraightquotes{image\_id}: 32677, \doublestraightquotes{question}: \doublestraightquotes{What is in the image?}, \doublestraightquotes{answer}: \doublestraightquotes{A dog and a cat.}\}\\
            Explanation: ``What is in the image?'' may be applied for any image, and thus it is an undesired question.\\
            Example 2:\\
            Input: \{\doublestraightquotes{image\_id}: 32678, \doublestraightquotes{caption}: \doublestraightquotes{A yellow happy emoji.}\}\\
            Output: \{\doublestraightquotes{image\_id}: 32678, \doublestraightquotes{question}: \doublestraightquotes{What emotion does this emoji express?}, \doublestraightquotes{answer}: \doublestraightquotes{Happiness.}\}\\
            Undesired output: \{\doublestraightquotes{image\_id}: 32678, \doublestraightquotes{question}: \doublestraightquotes{What emotion does this image express?}, \doublestraightquotes{answer}: \doublestraightquotes{Happiness.}\}\\
            Explanation: Mentioning a specific object, emoji, implies that there must be an emoji in the image. On the other end, ``What emotion does this image express?'' may be applied for any image (one may say any image conveys some emotion).\\
            Example 3:\\
            Input: \{\doublestraightquotes{image\_id}: 34512, \doublestraightquotes{caption}:  \doublestraightquotes{An image of the Empire State Building.}\}\\
            Output: \{\doublestraightquotes{image\_id}: 34512, \doublestraightquotes{question}:  \doublestraightquotes{What is the name of the building in the image?}, \doublestraightquotes{answer}:  \doublestraightquotes{The Empire State Building.}\}\\
            Undesired output: \{\doublestraightquotes{image\_id}: 34512, \doublestraightquotes{question}: \doublestraightquotes{What place is it in the image?}, \doublestraightquotes{answer}:  \doublestraightquotes{The Empire State Building.}\}\\
            Explanation: Mentioning a specific object, building, implies that there must be a building in the image. On the other end, ``What place is it in the image?'' may be applied for almost any image.\\
            \end{minipage}
        }
        \caption{}
        \label{fig:supp:gpt_instructions:ivqd}
    \end{subfigure}
    \caption{The instructions given to GPT-4o mini for (a) generating incorrect answer options for AAD multiple-choice questions and (b) generating image-specific questions for IVQD multiple-choice questions.}
    \label{fig:supp:gpt_instructions}  
\end{figure*}

\section{Training datasets}
\label{supp:sec:training_datasets}

\subsection{Multiple-choice training dataset}
\label{supp:sec:training_datasets:multiple_choice}

We provide details about the dataset we created for training CLIP-UP on multiple-choice VQA. 
The goal was to create a compact high-quality UPD training dataset. We do not use the fine-tuning dataset from~\cite{miyai2025unsolvable} as it is too large (10,000 samples), lacks IASD samples, and, upon our manual inspection, found to be of insufficient quality.

The dataset is organized into multiple-choice VQA question pairs, each consisting of an answerable question and its corresponding unanswerable variant.
The training set contains 263, 159, and 277 question pairs for AAD, IASD, and IVQD, respectively (a total of 526, 318, and 554 samples). 
The validation set contains 30 pairs for each category. 
We do not include a test set.

Unlike the training set, each question in the validation set is augmented with $n$ repetitions ($n$ is the number of options), each with a different circular shift of the options, enabling dual accuracy evaluation.
Consequently, the validation set contains a total of 204, 232, and 226 questions for AAD, IASD, and IVQD, respectively.

Data were created with a different process for each unanswerability category, as we explain below.
All data were sourced from public training sets to ensure no leakage with public benchmarks and test sets.
For all categories, questions were generated with four options. Most questions were left unchanged, but some were modified to include fewer options. We also ensured that the correct option varies (\eg, it is not always ``A'').

Note that the dataset was constructed in a straightforward manner, resulting in structurally 
uniform questions, as we assumed this would suffice for training CLIP-UP.
This simplicity highlights CLIP-UP's robustness and suggests that a more diverse dataset could further improve performance.

\subsubsection{AAD data}

The AAD data consist of 293 pairs of questions: 143 sourced from the A-OKVQA dataset~\cite{schwenk2022okvqa}, and 150 generated using GPT-4o mini~\cite{openai2024gpt4o} based on MS COCO~\cite{lin2014microsoft}. 

Our goal is to have standard questions with exactly one correct answer option, while all others are clearly incorrect. This ensures that AAD unanswerable questions may be generated by removing the correct answer option, leaving no valid answer in the answer options set. 
Note that this condition is not always met, as many multiple-choice questions are intentionally designed to be challenging, requiring the selection of the best option from several plausible ones.

We began by creating the standard questions, selecting 143 multiple-choice VQA items from the A-OKVQA training dataset~\cite{schwenk2022okvqa}. 
We manually examined the data to include only questions with exactly one correct answer option.

We created 150 additional standard questions using the following process: we first sampled examples from MS COCO training set~\cite{lin2014microsoft} (2017 split). Each sample consists of an image and five ground truth captions, from which we randomly selected one. 
Next, we used GPT-4o mini~\cite{openai2024gpt4o} to generate three incorrect captions for each sample. GPT-4o mini was given an image and its correct caption, and instructed to output a multiple-choice VQA question asking to select the correct caption, with four answer options: a correct one (the ground truth caption) and three incorrect ones (generated by GPT-4o mini). See the instruction used in \cref{fig:supp:gpt_instructions:aad}. 
To diversify the data, we alternated between two question formats: ``Which caption describes the image?'' and ``Which one is the correct caption for this image?''. As with the A-OKVQA questions, we included only standard questions with exactly one correct answer option.

After obtaining 293 standard multiple-choice VQA questions from both sources, we created the AAD counterparts by removing the correct answer option from each standard question. See \cref{fig:supp:dataset_examples:AAD} for an example.

\subsubsection{IASD data}

The IASD data consist of 189 pairs of questions. 
In the case of IASD, there are no specific constraints on the standard questions.
However, for unanswerable questions, the textual question (the question itself, \eg, ``What color is the dress?'') and the answer options set must be incompatible.

Similar to the AAD case, we used standard questions from the A-OKVQA training dataset~\cite{schwenk2022okvqa} and ones generated with GPT-4o mini~\cite{openai2024gpt4o}. To create the unanswerable counterpart for each standard question, the original answer set was replaced with one from another randomly selected standard question. 
We then manually examined the data to include only pairs where the textual question is genuinely incompatible with the unanswerable answer options set. See \cref{fig:supp:dataset_examples:IASD} for an example.

\subsubsection{IVQD data}

The IVQD data consist of 307 pairs of questions: 42 sourced from the fine-tuning data by~\cite{miyai2025unsolvable}, and 265 generated using GPT-4o mini~\cite{openai2024gpt4o} based on MS COCO~\cite{lin2014microsoft} and TextCaps~\cite{sidorov2020textcaps}.

Our goal is to have pairs of multiple-choice VQA questions where the textual question conveys some specific information about the image.
This allows generating unanswerable IVQD questions by replacing the image with another image that is incompatible with the information in the textual question (in contrast, non-specific questions like ``What emotion does this image convey?'' are compatible with most images).

The 42 pairs sourced from the fine-tuning data by~\cite{miyai2025unsolvable} include corresponding standard and IVQD unanswerable questions. 
We manually ensured that in all pairs, the textual question conveys image-specific information and is genuinely incompatible with the image in the unanswerable item.

For the 265 other question pairs, we generated standard questions using the following process: similar to the AAD case, we sampled examples from MS COCO training set~\cite{lin2014microsoft} (2017 split), but also from TextCaps training set~\cite{sidorov2020textcaps}. Each sample consists of an image and five ground truth captions, from which we randomly selected one. 
Next, we used GPT-4o mini~\cite{openai2024gpt4o} to generate an image-specific textual question from each caption. GPT-4o mini was given a caption (without the image) and instructed to output an image-specific textual question related to the caption along with the correct answer. See \cref{fig:supp:gpt_instructions:ivqd} for the instruction used.
Then, we used GPT-4o mini to create three incorrect answer options for each question by providing it with the image, question and correct answer as input, and instructing it similarly to the AAD case.

To create the unanswerable counterpart for each standard question, we replaced the image with one from another randomly selected standard question. The data were manually reviewed to include only pairs where the textual question is 
image-specific and genuinely incompatible with unanswerable IVQD image. See \cref{fig:supp:dataset_examples:IVQD} for an example.

\begin{table*}[!t]
    \centering
    \setlength{\tabcolsep}{2.5pt}
        \begin{tabular}{c|ccc|ccc|ccc|ccc}
        \toprule
        \multirow{2}{*}{\textbf{Method}} & \multicolumn{3}{c|}{\textbf{AAD}} & \multicolumn{3}{c|}{\textbf{IASD}} & \multicolumn{3}{c|}{\textbf{IVQD}} & \multicolumn{3}{c}{\textbf{Average}} \\
        & Stand. & UPD & Dual 
        & Stand. & UPD & Dual 
        & Stand. & UPD & Dual
        & Stand. & UPD & Dual \\
        \midrule
        \multicolumn{1}{l|}{Base Setting} & 
        69.02 & 1.71 & 1.59 &
        66.49 & 19.70 & 12.73 & 
        63.48 & 0.28 & 0.28 & 
        66.33 & 7.23 & 4.87 \\
        
        \multicolumn{1}{l|}{CLIP-UP-Emb ($\alpha = 0.0$)} & 
         69.02 & 1.46 & 1.34 &
         66.27 & 19.59 & 12.84 &
         63.76 & 0.56 & 0.56 & 
         66.35 & 7.20 & 4.91 \\

        \multicolumn{1}{l|}{CLIP-UP-Emb ($\alpha = 0.1$)} & 
         68.90 & 1.59 & 1.46 &
         66.70 & 19.48 & 12.95 &
         63.48 & 0.28 & 0.28 & 
         66.36 & 7.12 & 4.90 \\

        \multicolumn{1}{l|}{CLIP-UP-Emb ($\alpha = 0.2$)} & 
        69.15 & 1.46 & 1.46 &
        66.59 & 20.13 & 13.28 &
        62.92 & 0.28 & 0.28 & 
        66.22 & 7.29 & 5.01 \\

        \multicolumn{1}{l|}{CLIP-UP-Emb ($\alpha = 0.3$)} & 
         69.15 & 1.59 & 1.46 &
         66.92 & 20.35 & 13.49 &
         63.20 & 0.56 & 0.28 & 
         66.42 & 7.50 & 5.08 \\
         
        \multicolumn{1}{l|}{CLIP-UP-Emb ($\alpha = 0.4$)} & 
         69.27 & 1.22 & 1.22 &
         66.81 & 19.15 & 12.51 &
         64.04 & 0.56 & 0.28 & 
         66.71 & 6.98 & 4.67 \\

        \multicolumn{1}{l|}{CLIP-UP-Emb ($\alpha = 0.5$)} & 
         69.39 & 0.73 & 0.73 &
         67.03 & 12.19 & 7.94 &
         64.04 & 0.56 & 0.28 & 
         66.82 & 4.49 & 2.98 \\
         
        \multicolumn{1}{l|}{CLIP-UP-Emb ($\alpha = 0.6$)} & 
         67.68 & 36.95 & 33.17 &
         66.16 & 45.05 & 30.90 &
         62.64 & 32.58 & 22.19 & 
         65.49 & 38.19 & 28.75 \\

        \multicolumn{1}{l|}{CLIP-UP-Emb ($\alpha = 0.7$)} & 
         63.78 & 57.93 & 45.00 &
         62.46 & 80.41 & 51.80 &
         60.67 & 68.82 & 45.51 & 
         62.30 & 69.05 & 47.44 \\  

        \multicolumn{1}{l|}{CLIP-UP-Emb ($\alpha = 0.8$)} & 
         61.71 & 65.00 & 47.07 &
         59.74 & 88.25 & 53.75 &
         58.99 & 78.65 & 49.44 & 
         60.15 & 77.30 & 50.09 \\

        \multicolumn{1}{l|}{CLIP-UP-Emb ($\alpha = 0.9$)} & 
         61.46 & 67.07 & 47.68 &
         59.09 & 90.75 & 54.19 &
         58.15 & 83.15 & 51.40 & 
         59.57 & 80.32 & 51.09 \\

        \multicolumn{1}{l|}{Original CLIP-UP-Emb ($\alpha = 1.0$)} & 
        61.22 & 67.68 & 47.80 &  
        59.52 & 90.64 & 54.73 & 
        58.43 & 82.58 & 51.12 & 
        59.72 & 80.30 & 51.22 \\
        
        \bottomrule
    \end{tabular}
    \caption{Standard-dual accuracy trade-off control results on LLaVA-1.5-7B.}
    \label{table:inference_time_control}
\end{table*}

\subsection{Open-ended training dataset}
\label{supp:sec:training_datasets:open_ended}

The open-ended VQA training set consists of 700 corresponding answerable-unanswerable question pairs sampled from the TDIUC training set~\cite{kafle2017analysis}, with 50 pairs allocated for validation.
Unanswerable questions were first drawn from the ``Absurd'' category. Each question was then paired with an answerable question from a valid (non-absurd) category about the same image.

\section{Evaluation details}
\label{supp:sec:evaluation}

\subsection{Multiple-choice baselines}
\label{supp:subsec:multiple_choice_baselines}

We provide details on the three prompt engineering settings from~\cite{miyai2025unsolvable}:
(1)~Base Prompt Setting: uses only the multiple-choice VQA prompt without additional instructions. As this setting does not explicitly encourage choosing an answer, it identifies unanswerable questions better than the original setup;
(2)~Additional-Option Setting: adds an option depending on the unanswerability category (``None of the above'' for AAD and IASD, or ``The image and question are irrelevant'' for IVQD), and includes the original instruction. This setting ensures that a correct answer is always present and encourages the model to select one;
(3)~Additional-Instruction Setting: adds an instruction to encourage withholding an answer when appropriate. The instructions vary by the unanswerability category and are similar to the extra option in the Additional-Option Setting.

Note that settings (2) and (3) assume knowledge of the input’s unanswerability category, which is not the case in real-world scenarios. They are thus meant to test models' capabilities via prompt engineering rather than serve as practical solutions.

\subsection{Multiple-choice evaluation}
\label{supp:subsec:multiple_choice_evaluation}

We conducted all the multiple-choice UPD evaluations ourselves, including those of the prompt engineering methods. 
For all experiments that were also performed by~\citet{miyai2025unsolvable}, our results closely align with theirs.

Multiple-choice UPD evaluation requires extracting the selected option from the model's prediction. We followed the extraction approach described in~\cite{miyai2025unsolvable}: 
each VLM prediction is first processed using a string matching algorithm, and if this fails, GPT-3.5 (gpt-3.5-turbo-0125~\cite{openai2023gpt35turbo0125}) is employed with a tailored prompt to extract the selected option.
We introduced slight modifications to the string matching algorithm to improve efficiency and accuracy, and reduce calls to GPT-3.5. To ensure a fair comparison, all results were evaluated using our modified string matching extraction algorithm.

\subsection{Open-ended evaluation}
\label{supp:subsec:open_ended_evaluation}

We evaluate on two open-ended VQA test datasets.
The first is sampled from the RGQA benchmark~\cite{zhang2023toward} and consists of 2{,}500 pairs of corresponding answerable-unanswerable questions. 
Pairs were randomly drawn from all four RGQA subsets (CLIP-easy, PT-easy, CLIP-hard, and PT-hard), with 625 pairs from each subset. 
Each pair shares the same image but contains a different question.

The second dataset consists of 2{,}500 pairs of corresponding answerable-unanswerable questions sampled from the TDIUC benchmark~\cite{kafle2017analysis}.  
The data was collected similarly to the open-ended training set, with the key difference that here the samples were drawn from the TDIUC test set.

We adopt the LAVE evaluation metric~\cite{manas2024improving}, which leverages an LLM for open-ended VQA scoring and has been shown to align better with human judgment than alternative metrics.
We use GPT-3.5 (gpt-3.5-turbo-0125~\cite{openai2023gpt35turbo0125}) as the LLM.

\section{Parsing and classification of question prompts}
\label{supp:sec:parsing}

This section describes the rule-based algorithm mentioned in the main paper. 
The algorithm serves two purposes: first, to classify whether a textual input is a multiple-choice question, an open-ended question, or neither. 
This classification determines whether, and which, correlation vector should be generated and integrated into the VLM. 
It can be applied to InjLoRA, and even to standard LoRA fine-tuned models, to decide whether LoRA weights should be used (as long as LoRA weights are not merged into the base model).

Second, if the input is a multiple-choice question, the algorithm parses it to separate the textual question and answer options, a step necessary for generating the correlation vectors.

The algorithm relies on simple string matching and assumes a specific structure of multiple-choice question prompts: a question followed by answer options, each preceded by a letter (\eg, ``A''). For example, ``What animal is by the flowers? A. Dog B. Rabbit C. Cat.''

In the first step, the algorithm checks whether the input is a multiple-choice question by detecting for the presence of ``A.'' and ``B.'' (since a question must have at least two options). 
If these are present in the input, the algorithm proceeds to the next step, where it parses the input: the question is the text before ``A.'', the first answer option is the text between ``A.'' and ``B.'', and so on for the remaining answer options. 
If the input is not classified as multiple-choice, the algorithm simply checks for the presence of a question mark to determine whether it is an open-ended question (or not a question at all).

Since the algorithm relies on string matching, it can be easily adjusted to support different multiple-choice input formats (\eg, options denoted with numbers instead of letters).
Moreover, the algorithm could easily be replaced with a more sophisticated approach, such as leveraging the LLM component of the VLM for more robust detection.
We however found it unnecessary given the simplicity of the parsing task on our test data.

\begin{table*}[!t]
    \centering
    \setlength{\tabcolsep}{2.5pt} 
    \begin{subtable}{\linewidth}
        \centering

        \begin{tabular}{c|ccc|ccc|ccc|ccc}
            \toprule
            \multirow{2}{*}{\textbf{Method}} & \multicolumn{3}{c|}{\textbf{AAD}} & \multicolumn{3}{c|}{\textbf{IASD}} & \multicolumn{3}{c|}{\textbf{IVQD}} & \multicolumn{3}{c}{\textbf{Average}} \\
            & Stand. & UPD & Dual 
            & Stand. & UPD & Dual 
            & Stand. & UPD & Dual
            & Stand. & UPD & Dual \\
            \midrule
            \multicolumn{1}{l|}{Original Model} & 
            70.24 & 0.00 & 0.00 & 
            67.79 & 0.33 & 0.33 & 
            65.17 & 0.00 & 0.00 & 
            67.73 & 0.11 & 0.11 \\

            \multicolumn{1}{l|}{Base Setting} & 
            69.02 & 1.71 & 1.59 &
            66.49 & 19.70 & 12.73 & 
            63.48 & 0.28 & 0.28 & 
            66.33 & 7.23 & 4.87 \\

            \multicolumn{1}{l|}{Additional-Option} & 
            68.41 & 48.54 & 40.85 &
            65.72 & 78.24 & 51.58 & 
            63.76 & 26.97 & 21.91 & 
            65.96 & 51.25 & 38.11 \\

            \multicolumn{1}{l|}{Additional-Instruction} & 
            68.54 & 33.90 & 27.80 &
            65.61 & 65.18 & 42.76 &
            63.48 & 31.74 & 23.60 & 
            65.88 & 43.61 & 31.39 \\
            
            \multicolumn{1}{l|}{Correlation Vectors Classifier} & 
             30.24 & 99.63 & 30.00 &
             29.16 & 63.87 & 17.30 &
             46.08 & 90.59 & 28.65 & 
             35.16 & 84.70 & 25.32 \\

            \multicolumn{1}{l|}{LoRA Fine-Tuning} & 
            64.63 & 56.34 & 43.78 &
            61.92 & 87.81 & 54.73 &
            61.24 & 85.96 & 52.25 & 
            62.60 & 76.70 & 50.25 \\
            
            \multicolumn{1}{l|}{LoRA Fine-Tuning (CLIP-UP data)} & 
            66.71 & 51.10 & 40.49 &
            63.33 & 88.25 & 55.71 & 
            61.80 & 67.70 & 44.66 & 
            63.95 & 69.02 & 46.95 \\

            \multicolumn{1}{l|}{CLIP-UP-Emb (ours)} &
            61.22 & 67.68 & 47.80 &  
            59.52 & 90.64 & 54.73 & 
            58.43 & 82.58 & 51.12 & 
            59.72 & 80.30 & 51.22 \\
    
            \multicolumn{1}{l|}{CLIP-UP-EmbLoRA (ours)} & 
            62.44 & 71.71 & \textbf{49.27} & 
            60.17 & 93.47 & \textbf{56.58} & 
            59.55 & 84.83 & \textbf{53.37} & 
            60.72 & 83.34 & \textbf{53.07} \\ 
  
            \bottomrule
        \end{tabular}
        \caption{LLaVA-1.5-7B}
        \label{table:full_mm_upd_results1_supp:llava_1.5_7b_results}
    \end{subtable}
    \vfill
    \vspace{0.3cm}
    \begin{subtable}{\linewidth}
        \centering
        \begin{tabular}{c|ccc|ccc|ccc|ccc}
            \toprule
            \multirow{2}{*}{\textbf{Method}} & \multicolumn{3}{c|}{\textbf{AAD}} & \multicolumn{3}{c|}{\textbf{IASD}} & \multicolumn{3}{c|}{\textbf{IVQD}} & \multicolumn{3}{c}{\textbf{Average}} \\
            & Stand. & UPD & Dual 
            & Stand. & UPD & Dual 
            & Stand. & UPD & Dual
            & Stand. & UPD & Dual \\
            \midrule
            \multicolumn{1}{l|}{Original Model} & 
            76.71 & 0.00 & 0.00 & 
            73.23 & 0.11 & 0.00 & 
            71.35 & 0.00 & 0.00 &
            73.76 & 0.04 & 0.00 \\
            
            \multicolumn{1}{l|}{Base Setting} & 
            72.32 & 23.78 & 17.80 &
            69.75 & 49.62 & 31.66 & 
            68.82 & 44.66 & 33.15 & 
            70.30 & 39.35 & 27.54 \\
            
            \multicolumn{1}{l|}{Additional-Option} & 
            75.85 & 18.41 & 18.05 &
            72.47 & 39.28 & 29.92 & 
            70.79 & 46.35 & 38.20 & 
            73.04 & 34.68 & 28.72 \\

            \multicolumn{1}{l|}{Additional-Instruction} & 
            67.07 & 48.66 & 38.29 &
            63.87 & 87.81 & 57.02 &
            68.82 & 71.91 & 54.49 & 
            66.59 & 69.46 & 49.93 \\
            
            \multicolumn{1}{l|}{Chain-of-Thought} &
            60.00 & 60.50 & 42.80 & 
            56.40 & 70.80 & 43.90 & 
            59.00 & 75.30 & 47.50 & 
            58.47 & 68.87 & 44.73 \\
            
            \multicolumn{1}{l|}{Self-Reflection} &
            66.20 & 50.00 & 37.80 & 
            62.60 & 55.80 & 36.70 &
            59.80 & 61.50 & 39.00 & 
            62.87 & 55.77 & 37.83 \\
            
            \multicolumn{1}{l|}{LoRA Fine-Tuning} & 
            69.15 & 58.54 & 47.56 &
            65.51 & 91.19 & 59.85 &      
            67.42 & 86.24 & \textbf{59.55} & 
            67.36 & 78.66 & 55.65 \\
            
            \multicolumn{1}{l|}{CLIP-UP-Emb (ours)} & 
            62.07 & 83.90 & 54.02 &
            58.54 & 95.65 & 55.71 & 
            57.87 & 92.70 & 55.06 & 
            59.49 & 90.75 & 54.93 \\
            
            \multicolumn{1}{l|}{CLIP-UP-EmbLoRA (ours)} &
            67.07 & 75.73 & \textbf{54.27} & 
            63.22 & 95.76 & \textbf{60.50} &
            63.48 & 84.83 & 53.65 & 
            64.59 & 85.44 & \textbf{56.14} \\
            \bottomrule
        \end{tabular}
        \caption{LLaVA-NeXT-13B}
        \label{table:full_supp_results:llava_next_13b_results}
    \end{subtable}
    \vfill
    \vspace{0.3cm}
    \begin{subtable}{\linewidth}
        \centering
        \begin{tabular}{c|ccc|ccc|ccc|ccc}
            \toprule
            \multirow{2}{*}{\textbf{Method}} & \multicolumn{3}{c|}{\textbf{AAD}} & \multicolumn{3}{c|}{\textbf{IASD}} & \multicolumn{3}{c|}{\textbf{IVQD}} & \multicolumn{3}{c}{\textbf{Average}} \\
            & Stand. & UPD & Dual 
            & Stand. & UPD & Dual 
            & Stand. & UPD & Dual
            & Stand. & UPD & Dual \\
            \midrule
            
            \multicolumn{1}{l|}{Original Model} & 
            80.73 & 1.22 & 1.22 & 
            77.48 & 0.00 & 0.00 & 
            77.53 & 0.00 & 0.00 & 
            78.58 & 0.41 & 0.41 \\
            
            \multicolumn{1}{l|}{Base Setting} & 
            79.51 & 1.95 & 1.95 &
            76.28 & 0.54 & 0.44 & 
            75.56 & 0.28 & 0.28 & 
            77.12 & 0.92 & 0.89 \\ 
            
            \multicolumn{1}{l|}{Additional-Option} & 
            80.24 & 23.41 & 21.95 &
            77.04 & 31.56 & 24.27 & 
            75.28 & 64.33 & 51.97 & 
            77.52 & 39.77 & 32.73 \\ 
            
            \multicolumn{1}{l|}{Additional-Instruction} & 
            77.93 & 31.95 & 27.93 &
            74.76 & 46.25 & 32.86 &
            74.72 & 72.75 & 56.18 & 
            75.80 & 50.32 & 38.99 \\ 
            
            \multicolumn{1}{l|}{LoRA Fine-Tuning} & 
            61.83 & 71.95 & 47.93 &
            59.52 & 95.21 & 56.69 &      
            58.71 & 93.82 & 56.46 & 
            60.02 & 86.99 & 53.69 \\
            
            \multicolumn{1}{l|}{CLIP-UP-Emb (ours)} & 
            62.68 & 81.95 & 52.80 &
            59.52 & 88.25 & 52.67 &
            60.67 & 93.26 & 56.74 & 
            60.96 & 87.82& 54.07 \\
            
            \multicolumn{1}{l|}{CLIP-UP-EmbLoRA (ours)} & 
            72.32 & 77.44 & \textbf{60.61} & 
            69.10 & 91.40 & \textbf{63.00} &
            69.10 & 91.01 & \textbf{63.20} & 
            70.17 & 86.62 & \textbf{62.27} \\

            \bottomrule
        \end{tabular}
        \caption{Phi-3.5-Vision}
        \label{table:full_supp_results:phi_vision_results}
    \end{subtable}

    \caption{
    Full results (\%) on MM-UPD~\cite{miyai2025unsolvable} multiple-choice VQA for (a) LLaVA-1.5-7B, (b) LLaVA-NeXT-13B, and (c) Phi-3.5-Vision. 
    Metrics include circular standard, UPD, and dual accuracies.}
    \label{table:full_mm_upd_results1_supp}
\end{table*}

\section{Standard-dual accuracy trade-off control}
\label{supp:sec:inference_time_control}

Although CLIP-UP enhances VLMs' UPD capabilities, reflected in improved dual accuracy, this comes at the cost of reduced standard accuracy, introducing a trade-off between the two.
In some applications, particularly those requiring high reliability (\eg, medical VQA systems), a steep drop in standard accuracy may be unacceptable. This highlights the need for a mechanism to control this trade-off.

We introduce a simple inference-time method, requiring no retraining, for controlling the trade-off. This is done by interpolating CLIP-UP-Emb's embedding vector ($\mathbf{e}$ from~\cref{eq:clip_up_projection}) with random noise:

\begin{equation}
\label{eq:inference_time_control}
    \mathbf{e} = \alpha \mathbf{e} + (1 - \alpha) \mathbf{e}_{noise},
\end{equation} 

where $\alpha \in [0,1]$ controls the interpolation strength and $\mathbf{e}_{noise}$ is sampled from the standard normal distribution.

\cref{table:inference_time_control} shows the MM-UPD results on LLaVA-1.5-7B.
First, using only noise ($\alpha = 0.0$) performs similarly to the base setting, thus not altering model behavior, motivating the idea of noise interpolation.
As $\alpha$ increases, dual accuracy improves while standard accuracy decreases, reflecting a controllable trade-off.
For some $\alpha$ values, adjusting CLIP-UP for higher standard accuracy still preserves strong dual performance. 
For example, $\alpha = 0.7$ yields a $2.58\%$ gain in standard accuracy with only a $3.78\%$ drop in dual accuracy.

\section{Additional results}
\label{supp:sec:additional_experiments}

\subsection{Full MM-UPD results}
\label{supp:subsec:full_mm_upd_results}

\begin{figure*}[!t]
\small
\centering
\begin{tabular}{L{0.16\textwidth}|L{0.37\textwidth}|L{0.37\textwidth}}
\toprule
 & \multicolumn{1}{c|}{\textbf{Standard}} & \multicolumn{1}{c}{\textbf{Unanswerable}} \\
\midrule

\multirow{3}{=}{\makecell{\textbf{Question}}} 
& \multicolumn{2}{c}{%
    \includegraphics[width=0.30\textwidth]{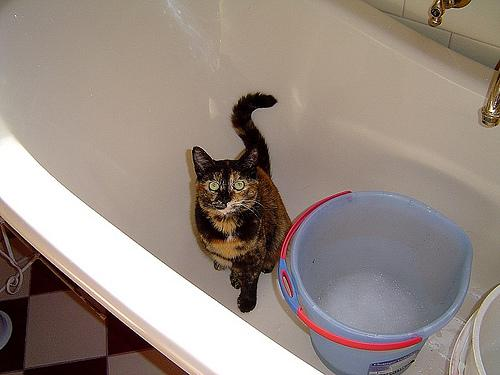}
} \\
& \parbox[t]{0.37\textwidth}{How many cats are visible in this picture?\\A. Three\\B. Four\\C. Two\\D. One} 
& \parbox[t]{0.37\textwidth}{How many cats are visible in this picture?\\A. Three\\B. Four\\C. Two} \\
\midrule

LLaVA-1.5-7B 
& D. \cmark 
& C. two. \xmark \\
\graymidrule
LLaVA-1.5-7B + CLIP-UP-EmbLoRA
& D. one. \cmark 
& I cannot answer. \cmark \\
\midrule

LLaVA-NeXT-13B
& D. \cmark 
& C. two. \xmark \\
\graymidrule
LLaVA-NeXT-13B + CLIP-UP-EmbLoRA
& D. one. \cmark 
& I cannot answer. \cmark \\
\midrule

Phi-3.5-Vision
& D. \cmark 
& C. \xmark \\
\graymidrule
Phi-3.5-Vision + CLIP-UP-EmbLoRA
& D. \cmark 
& I cannot answer. \cmark \\
\midrule

InternVL3-1B
& D. one. \cmark 
& C. two. \xmark \\
\graymidrule
InternVL3-1B + CLIP-UP-EmbLoRA
& D. one. \cmark 
& C. two. \xmark \\
\midrule

InternVL3-8B
& D. one. \cmark 
& C. two. \xmark \\
\graymidrule
InternVL3-8B + CLIP-UP-EmbLoRA
& D. one. \cmark 
& I cannot answer. \cmark \\
\midrule

Ovis2-16B
& D. one. \cmark 
& In the image, there is only one cat visible, sitting in a bathtub. Therefore, the correct answer is C. two, as it is the closest option to the actual number of cats present. \xmark \\
\graymidrule
Ovis2-16B + CLIP-UP-EmbLoRA
& D. one. \cmark
& I cannot answer. \cmark \\
\midrule

\end{tabular}
\caption{Models predictions on a pair of standard and AAD unanswerable multiple-choice VQA questions from MM-UPD~\cite{miyai2025unsolvable}, without and with CLIP-UP-EmbLoRA.}
\label{fig:mm_upd_aad_performance_example}
\end{figure*}

\cref{table:full_mm_upd_results1_supp,table:full_mm_upd_results2_supp} present the complete multiple-choice MM-UPD results, including LLaVA-NeXT-13B and InternVL3-1B, as well as results for each unanswerability category (AAD, IASD, and IVQD).
\cref{fig:mm_upd_ivqd_performance_example,fig:mm_upd_aad_performance_example} show VLM responses to answerable and unanswerable questions, with and without CLIP-UP.

\begin{table*}[!t]
    \centering
    \setlength{\tabcolsep}{2.5pt}
    \begin{subtable}{\linewidth}
        \centering

        \begin{tabular}{c|ccc|ccc|ccc|ccc}
            \toprule
            \multirow{2}{*}{\textbf{Method}} & \multicolumn{3}{c|}{\textbf{AAD}} & \multicolumn{3}{c|}{\textbf{IASD}} & \multicolumn{3}{c|}{\textbf{IVQD}} & \multicolumn{3}{c}{\textbf{Average}} \\
            & Stand. & UPD & Dual 
            & Stand. & UPD & Dual 
            & Stand. & UPD & Dual
            & Stand. & UPD & Dual \\
            \midrule
            \multicolumn{1}{l|}{Original Model} & 
            76.59 & 0.00 & 0.00 & 
            74.10 & 0.00 & 0.00 & 
            74.44 & 0.00 & 0.00 & 
            75.04 & 0.00 & 0.00 \\
            
            \multicolumn{1}{l|}{Base Setting} & 
            71.22 & 5.85 & 5.37 & 
            67.03 & 6.53 & 2.50 & 
            67.98 & 2.53 & 1.69 & 
            68.74 & 4.97 & 3.19 \\
            
            \multicolumn{1}{l|}{Additional-Option} & 
            75.49 & 36.83 & 35.73 & 
            73.01 & 41.35 & 30.79 & 
            74.44 & 26.40 & 20.79 & 
            74.31 &	34.86 &	29.10 \\
            
            \multicolumn{1}{l|}{Additional-Instruction} & 
            74.02 & 0.12 & 0.12 & 
            71.93 & 1.96 & 1.09 & 
            72.19 & 0.00 & 0.00 & 
            72.71 & 0.69 & 0.40 \\
            
            \multicolumn{1}{l|}{LoRA Fine-Tuning} & 
            69.27 & 62.20 & 51.10 & 
            67.25 & 85.96 & 58.43 & 
            66.29 & 80.62 & 53.09 & 
            67.60 & 76.26 & 54.21 \\
            
            \multicolumn{1}{l|}{CLIP-UP-Emb (ours)} & 
            64.63 & 64.02 & 46.22 & 
            60.39 & 76.71 & 47.12 & 
            61.80 & 89.89 & 56.18 & 
            62.27 &	76.87 & 49.84 \\
            
            \multicolumn{1}{l|}{CLIP-UP-EmbLoRA (ours)} &
            71.95 & 71.95 & \textbf{56.10} & 
            68.99 & 86.40 & \textbf{58.54} & 
            70.22 & 88.76 & \textbf{61.52} & 
            70.39 &	82.37 & \textbf{58.72} \\
    
            \bottomrule
        \end{tabular}
        \caption{InternVL3-1B}
        \label{table:full_supp_results:intern_vl3_1b}
    \end{subtable}
    \vfill
    \vspace{0.3cm}
    \begin{subtable}{\linewidth}
        \centering

        \begin{tabular}{c|ccc|ccc|ccc|ccc}
            \toprule
            \multirow{2}{*}{\textbf{Method}} & \multicolumn{3}{c|}{\textbf{AAD}} & \multicolumn{3}{c|}{\textbf{IASD}} & \multicolumn{3}{c|}{\textbf{IVQD}} & \multicolumn{3}{c}{\textbf{Average}} \\
            & Stand. & UPD & Dual 
            & Stand. & UPD & Dual 
            & Stand. & UPD & Dual
            & Stand. & UPD & Dual \\
            \midrule
            \multicolumn{1}{l|}{Original Model} & 
            91.10 & 0.00 & 0.00 & 
            87.27 & 0.00 & 0.00 & 
            88.48 & 0.00 & 0.00 & 
            88.95 & 0.00 & 0.00 \\
            
            \multicolumn{1}{l|}{Base Setting} & 
            88.17 & 37.07 & 35.73 & 
            84.00 & 52.77 & 45.81 & 
            86.80 & 43.54 & 36.80 & 
            86.32 & 44.46 & 39.45 \\
            
            \multicolumn{1}{l|}{Additional-Option} & 
            90.73 & 51.34 & 50.49 & 
            86.94 & 73.45 & 63.11 & 
            88.76 & 74.72 & 66.57 & 
            88.81 & 66.50 & 60.06 \\
            
            \multicolumn{1}{l|}{Additional-Instruction} & 
            87.93 & 59.88 & 56.95 & 
            84.87 & 92.38 & 78.24 & 
            86.52 & 88.20 & 76.12 & 
            86.44 & 80.15 & 70.44 \\
            
            \multicolumn{1}{l|}{LoRA Fine-Tuning} & 
            88.05 & 66.34 & 62.68 & 
            84.00 & 94.56 & 79.33 & 
            86.80 & 85.39 & 72.75 & 
            86.28 & 82.10 & 71.59 \\
            
            \multicolumn{1}{l|}{CLIP-UP-Emb (ours)} & 
            86.22 & 80.00 & 73.17 & 
            82.05 & 96.30 & 78.78 & 
            83.99 & 96.63 & 80.90 & 
            84.09 & 90.98 & 77.62 \\
            
            \multicolumn{1}{l|}{CLIP-UP-EmbLoRA (ours)} & 
            87.07 & 80.49 & \textbf{74.15} & 
            83.35 & 98.04 & \textbf{81.50} & 
            87.08 & 94.10 & \textbf{82.02} & 
            85.83 & 90.88 & \textbf{79.22} \\

            \bottomrule
        \end{tabular}
        \caption{InternVL3-8B}
        \label{table:full_supp_results:intern_vl3_8b}
    \end{subtable}

    \begin{subtable}{\linewidth}
        \centering

        \begin{tabular}{c|ccc|ccc|ccc|ccc}
            \toprule
            \multirow{2}{*}{\textbf{Method}} & \multicolumn{3}{c|}{\textbf{AAD}} & \multicolumn{3}{c|}{\textbf{IASD}} & \multicolumn{3}{c|}{\textbf{IVQD}} & \multicolumn{3}{c}{\textbf{Average}} \\
            & Stand. & UPD & Dual 
            & Stand. & UPD & Dual 
            & Stand. & UPD & Dual
            & Stand. & UPD & Dual \\
            \midrule
            \multicolumn{1}{l|}{Original Model} & 
             89.51 & 16.95 & 16.95 & 
             86.29 & 12.40 & 11.43 & 
             89.33 & 0.56 & 0.56 & 
             88.38 & 9.97 & 9.65 \\

            \multicolumn{1}{l|}{Base Setting} & 
             87.93 & 44.27 & 43.54 & 
             84.11 & 48.75 & 42.22 & 
             89.04 & 46.35 & 41.57 & 
             87.03 & 46.46 & 42.44 \\

            \multicolumn{1}{l|}{Additional-Option} & 
            89.27 & 53.78 & 52.20 & 
            85.64 & 72.36 & 62.68 & 
            87.64 & 86.52 & 75.84 & 
            87.52 & 70.89 & 63.57 \\
            
            \multicolumn{1}{l|}{Additional-Instruction} & 
            88.05 & 64.76 & 62.56 & 
            84.77 & 83.57 & 71.06 & 
            88.20 & 92.13 & 81.46 & 
            87.01 & 80.15 & 71.69 \\

            \multicolumn{1}{l|}{LoRA Fine-Tuning} & 
            86.71 & 71.59 & 66.83 & 
            83.57 & 95.76 & 79.98 & 
            87.36 & 95.51 & \textbf{83.15} & 
            85.88 & 87.62 & 76.65 \\

            \multicolumn{1}{l|}{CLIP-UP-Emb (ours)} & 
            85.12 & 79.63 & 72.56 & 
            81.39 & 96.95 & 78.89 & 
            83.99 & 98.31 & 82.30 & 
            83.50 & 91.63 & 77.92 \\

            \multicolumn{1}{l|}{CLIP-UP-EmbLoRA (ours)} & 
            88.54 & 78.54 & \textbf{75.12} & 
            84.87 & 98.80 & \textbf{83.79} & 
            89.04 & 93.26 & 82.87 & 
            87.48 & 90.20 & \textbf{80.59} \\

            \bottomrule
        \end{tabular}
        \caption{Ovis2-16B}
        \label{table:full_supp_results:ovis2_16b}
    \end{subtable}

    \caption{
    Full results (\%) on MM-UPD~\cite{miyai2025unsolvable} multiple-choice VQA for (a) InternVL3-1B, (b) InternVL3-8B, and (c) Ovis2-16B. 
    Metrics include circular standard, UPD, and dual accuracies.}
    \label{table:full_mm_upd_results2_supp}
\end{table*}

\subsection{Additional baselines}
\label{supp:subsec:additional_baselines}

We include three additional baselines in~\cref{table:full_mm_upd_results1_supp}.
First, we present results of LLaVA-1.5-7B LoRA fine-tuning following~\cite{miyai2025unsolvable}, using CLIP-UP's multiple-choice training data. 
Fine-tuning was conducted under the same settings as the original setup, but with 3 epochs instead of one, for a fair comparison with CLIP-UP methods. 
This setting achieves reasonable performance but is inferior to the CLIP-UP methods, and to the original LoRA fine-tuning setup that uses more data. 
This suggests that CLIP-UP methods are more data-efficient.

Second, we tested a simple classifier-based baseline: a logistic regression classifier trained on our correlation vectors determines whether the input is answerable or unanswerable.
If classified as unanswerable, the output is ``I cannot answer''; otherwise, the VLM generates the response.
The classifier itself labels only 46.27\% of circular standard questions as answerable, which already sets a dual accuracy upper bound that indicates that the baseline underperforms CLIP-UP methods on all VLMs.
For completeness, we still evaluate the full baseline, with results for LLaVA-1.5-7B shown in~\cref{table:full_mm_upd_results1_supp:llava_1.5_7b_results} (line 5). 
The baseline substantially underperforms CLIP-UP variants, suggesting that injection is necessary and that the correlation vectors alone do not capture all relevant information.

Finally, we compare CLIP-UP to two prompt engineering methods proposed in~\cite{miyai2025unsolvable}. 
The first employs zero-shot Chain-of-Thought~\cite{kojima2022large} reasoning by appending the phrase ``Let’s think step by step'' to the multiple-choice VQA prompt, encouraging the model to reason more carefully. 
The second uses self-reflection~\cite{kadavath2022language} by prompting the model to evaluate its own response.
Although both techniques improve performance, CLIP-UP methods outperform them by a significant margin.
The results for these methods are taken directly from~\cite{miyai2025unsolvable} and reported only for LLaVA-NeXT-13B.

\subsection{Additional open-ended results}
\label{supp:subsec:open_ended_results}

\cref{table:rgqa_open_ended_results_sup} presents the RGQA results for LLaVA-NeXT-13B, InternVL3-1B, and Ovis2-16B, not shown in the main paper.
\cref{fig:rgqa_performance_example} shows model responses to RGQA answerable and unanswerable questions, with and without CLIP-UP.

\cref{table:tdiuc_open_ended_results} presents results on open-ended questions from the TDIUC~\cite{kafle2017analysis} test set. 
CLIP-UP-EmbLoRA outperforms the baselines on three VLMs, while on most others the performance gap is small (under 0.6\%).
Notably, accuracies are high for all training-involved methods, as training was performed on samples from the TDIUC training set.

\begin{table*}[!t]
    \centering

    \begin{tabular}{c|ccc|ccc}
        \toprule
        \multirow{2}{*}{\textbf{Method}} &
        \multicolumn{3}{c|}{\textbf{LLaVA-NeXT-13B}} &
        \multicolumn{3}{c}{\textbf{InternVL3-1B}} \\
        & Stand. & UPD & Dual &
        Stand. & UPD & Dual \\
        \midrule
        \multicolumn{1}{l|}{Original Model} & 
        73.02 & 12.90 & 9.84 &
        64.40 & 11.10 & 7.46 \\
        
        \multicolumn{1}{l|}{Prompt Engineering} &
        68.32 & 33.76 & 22.64 &
        66.20 & 15.72 & 11.02 \\
        
        \multicolumn{1}{l|}{LoRA Fine-Tuning} &
        65.16 & 60.62 & 38.14 &
        56.68 & 60.04 & \textbf{31.10} \\
        
        \multicolumn{1}{l|}{CLIP-UP-Emb (ours)} &
        67.84 & 62.06 & \textbf{40.20} & 
        41.40 & 73.24 & 27.66 \\

        \multicolumn{1}{l|}{CLIP-UP-EmbLoRA (ours)} &
        55.26 & 75.94 & 39.42 &
        57.08 & 56.62 & 29.30 \\
        
        \bottomrule
    \end{tabular}
    \caption{Results (\%) on RGQA~\cite{zhang2023toward} open-ended VQA for LLaVA-NeXT-13B and InternVL3-1B. 
    Metrics include standard, UPD, and dual accuracies.}
    \label{table:rgqa_open_ended_results_sup}
\end{table*}

\begin{table*}[!t]
    \centering
    \begin{tabular}{c|ccc|ccc|ccc}
        \toprule
        \multirow{2}{*}{\textbf{Method}} &
        \multicolumn{3}{c|}{\textbf{LLaVA-1.5-7B}} &
        \multicolumn{3}{c|}{\textbf{LLaVA-NeXT-13B}} &
        \multicolumn{3}{c}{\textbf{Phi-3.5-Vision}} \\
        & Stand. & UPD & Dual &
        Stand. & UPD & Dual & 
        Stand. & UPD & Dual \\
        \midrule
        \multicolumn{1}{l|}{Original Model} & 
        84.84 & 10.50 & 9.10 & 
        87.70 & 11.66 & 10.28 &
        85.16 & 5.64 & 4.70 \\ 
        
        \multicolumn{1}{l|}{Prompt Engineering} &
        85.04 & 13.46 & 11.34 &
        86.18 & 73.46 & 63.44 &
        84.44 & 46.86 & 39.88 \\ 
        
        \multicolumn{1}{l|}{LoRA Fine-Tuning} &
        81.20 & 99.82 & 81.06 &
        84.62 & 99.96 & 84.58 &
        83.66 & 88.54 & 74.04 \\ 
        
        \multicolumn{1}{l|}{CLIP-UP-Emb (ours)} &
        81.42 & 98.46 & 80.24 &
        86.68 & 99.76 & 86.48 &
        82.38 & 98.98 & 81.64 \\
        
        \multicolumn{1}{l|}{CLIP-UP-EmbLoRA (ours)} &
        85.68 & 98.86 & \textbf{84.58} &
        87.00 & 99.58 & \textbf{86.58} &
        85.70 & 99.54 & \textbf{85.36} \\
 
        \bottomrule
    \end{tabular}
    \vfill
    \vspace{0.3cm}
    \begin{tabular}{c|ccc|ccc|ccc}
        \toprule
        \multirow{2}{*}{\textbf{Method}} &
        \multicolumn{3}{c|}{\textbf{InternVL3-1B}} &
        \multicolumn{3}{c|}{\textbf{InternVL3-8B}} &
        \multicolumn{3}{c}{\textbf{Ovis2-16B}} \\
        & Stand. & UPD & Dual 
        & Stand. & UPD & Dual 
        & Stand. & UPD & Dual \\
        \midrule
        \multicolumn{1}{l|}{Original Model} & 
        86.98 & 6.16 & 5.22 &
        89.06 & 14.04 & 12.42 &
        88.64 & 7.14 & 6.56 \\

        \multicolumn{1}{l|}{Prompt Engineering} &
        87.32 & 16.10 & 13.66  &
        84.70 & 95.28 & 80.80 &
        46.66 & 99.92 & 46.58 \\

        \multicolumn{1}{l|}{LoRA Fine-Tuning} &
        87.78 & 99.74 & \textbf{87.52} &
        89.54 & 99.82 & \textbf{89.36} &
        89.84 & 99.88 & \textbf{89.72} \\

        \multicolumn{1}{l|}{CLIP-UP-Emb (ours)} &
        83.58 & 99.24 & 82.94 &
        89.28 & 98.68 & 88.22 &
        88.52 & 99.74 & 88.26 \\

        \multicolumn{1}{l|}{CLIP-UP-EmbLoRA (ours)} &
        87.76 & 99.08 & 86.92 &
        89.80 & 99.42 & 89.30 &
        87.90 & 99.84 & 87.74 \\

        \bottomrule
    \end{tabular}
    \caption{Results (\%) on TDIUC~\cite{kafle2017analysis} open-ended VQA. 
    Metrics include standard, UPD, and dual accuracies.}
    \label{table:tdiuc_open_ended_results}
\end{table*}

\subsection{Ruling out potential RGQA bias}
\label{supp:subsec:rule_out_rgqa_bias}

We rule out the possibility of a CLIP-induced bias in the RGQA evaluation.
RGQA~\cite{zhang2023toward} comprises four subsets: CLIP-easy, CLIP-hard, PT-easy, and PT-hard, which differ in how unanswerable questions are generated.
In CLIP-easy, unanswerable questions are created by pairing texts and images with low CLIP similarity, while in CLIP-hard unanswerable questions are created from pairs with high CLIP similarity.
In contrast, PT-easy and PT-hard are constructed by modifying standard questions through random (PT-easy) or adversarial (PT-hard) word replacements.

Thus, VLMs equipped with CLIP-UP may benefit from questions in the CLIP-easy category. 
Although CLIP-hard is designed to have the opposite effect, the advantage on CLIP-easy may be more significant, potentially biasing our RGQA results in favor of CLIP-UP.

To examine this potential bias, we report in~\cref{table:pt_easy_pt_hard} the RGQA results on the PT-easy and PT-hard subsets only, which are not influenced by CLIP similarity.
We observe that evaluation trends are consistent across the full RGQA test set and the PT-only subsets (compare with~\cref{table:rgqa_open_ended_results_main,table:rgqa_open_ended_results_sup}): CLIP-UP-Emb or CLIP-UP-EmbLoRA outperform LoRA fine-tuning on all models except InternVL3-1B. This rules out the CLIP-easy bias concern.

\begin{table*}[!t]
    \centering
    \begin{tabular}{c|ccc|ccc|ccc}
        \toprule
        \multirow{2}{*}{\textbf{Method}} &
        \multicolumn{3}{c|}{\textbf{LLaVA-1.5-7B}} &
        \multicolumn{3}{c|}{\textbf{LLaVA-NeXT-13B}} &
        \multicolumn{3}{c}{\textbf{Phi-3.5-Vision}} \\
        & Stand. & UPD & Dual &
        Stand. & UPD & Dual & 
        Stand. & UPD & Dual \\
        \midrule
        \multicolumn{1}{l|}{Original Model} & 
        66.64 & 11.40 & 6.92 &
        71.56 & 12.68 & 9.28 &
        67.96 & 9.36 & 6.48 \\
        
        \multicolumn{1}{l|}{Prompt Engineering} &
        66.52 & 15.32 & 9.32 &
        66.88 & 35.60 & 22.76 &
        68.04 & 28.40 & 19.20 \\ 
        
        \multicolumn{1}{l|}{LoRA Fine-Tuning} &
        54.96 & 63.20 & 26.84 &
        64.24 & 59.72 & 34.84 &
        66.36 & 39.76 & 24.24 \\ 
        
        \multicolumn{1}{l|}{CLIP-UP-Emb (ours)} &
        36.60 & 76.44 & 23.52 &
        66.04 & 59.76 & \textbf{35.72} &
        58.24 & 57.92 & 27.04 \\
        
        \multicolumn{1}{l|}{CLIP-UP-EmbLoRA (ours)} &
        56.36 & 58.76 & \textbf{27.96} &
        53.16 & 74.04 & 34.72 &
        65.32 & 54.72 & \textbf{32.72} \\

        \bottomrule
    \end{tabular}
    \vfill
    \vspace{0.3cm}

    \begin{tabular}{c|ccc|ccc|ccc}
        \toprule
        \multirow{2}{*}{\textbf{Method}} &
        \multicolumn{3}{c|}{\textbf{InternVL3-1B}} &
        \multicolumn{3}{c|}{\textbf{InternVL3-8B}} &
        \multicolumn{3}{c}{\textbf{Ovis2-16B}} \\
        & Stand. & UPD & Dual &
        Stand. & UPD & Dual &
        Stand. & UPD & Dual \\
        \midrule
        \multicolumn{1}{l|}{Original Model} & 
        63.08 & 10.32 & 7.16 &
        69.72 & 13.52 & 10.12 &
        68.96 & 13.08 & 8.76 \\
        
        \multicolumn{1}{l|}{Prompt Engineering} &
        65.20 & 15.08 & 10.28 &
        67.00 & 47.12 & 29.40 &
        28.84 & 93.92 & 24.40 \\

        \multicolumn{1}{l|}{LoRA Fine-Tuning} &
        55.16 & 57.32 & \textbf{26.36} &
        65.56 & 56.80 & 33.44 &
        64.64 & 63.16 & 36.52 \\

        \multicolumn{1}{l|}{CLIP-UP-Emb (ours)} &
        39.32 & 70.64 & 22.32 &
        66.40 & 49.40 & 31.24 & 
        61.24 & 66.00 & 36.44 \\
        
        \multicolumn{1}{l|}{CLIP-UP-EmbLoRA (ours)} &
        55.32 & 54.88 & 25.48 &
        65.20 & 60.20 & \textbf{36.68} &
        62.12 & 69.08 & \textbf{38.72} \\

        \bottomrule
    \end{tabular}
    \caption{Results (\%) on PT-easy and PT-hard categories from RGQA~\cite{zhang2023toward} open-ended VQA. 
    Metrics include standard, UPD, and dual accuracies.}
    \label{table:pt_easy_pt_hard}
\end{table*}

\begin{table*}[!t]
    \centering
    \begin{tabular}{c|c|c|c|c|c|c}
        \toprule
        \multirow{1}{*}{\textbf{Method}} &
        \multicolumn{1}{c|}{\textbf{LLaVA-1.5}} &
        \multicolumn{1}{c|}{\textbf{LLaVA-Ne}} &
        \multicolumn{1}{c|}{\textbf{Phi-V-3.5}} &
        \multicolumn{1}{c|}{\textbf{InVL3-1B}} &
        \multicolumn{1}{c|}{\textbf{InVL3-8B}} &
        \multicolumn{1}{c}{\textbf{Ovis2-16B}} \\
        \midrule
        \multicolumn{1}{l|}{Original Model} & 
        57.30 & 64.60 & 61.70 & 62.50 & 72.50 & 73.50 \\

        \multicolumn{1}{l|}{CLIP-UP-Emb (ours)} &
        51.10 & 58.50 & 55.00 & 50.00 & 69.00 & 69.20 \\
        
        \multicolumn{1}{l|}{CLIP-UP-EmbLoRA (ours)} &
        51.20 & 60.50 & 59.20 & 60.00 & 71.30 & 70.90 \\
        \bottomrule
    \end{tabular}
    \caption{Results (\%) on multiple-choice answerable VQA from SEED-Bench~\cite{zhang2023toward}, reported with circular standard accuracy. 
    Evaluated models are LLaVA-1.5-7B, LLaVA-NeXT-13B, Phi-3.5-Vision, InternVL3-1B, InternVL3-8B, and Ovis2-16B.}
    \label{table:seed_bench}
\end{table*}

\subsection{Additional standard VQA evaluation}
\label{supp:subsec:seed_bench_eval}

We evaluated CLIP-UP on standard multiple-choice questions from SEED-Bench~\cite{li2023seed} to further assess standard accuracy drops.
We randomly sampled 1,000 examples from SEED-Bench's image categories and augmented each with circular duplicates, yielding 4,000 samples.
\cref{table:seed_bench} shows that CLIP-UP-EmbLoRA standard accuracy drops are minimal: up to 6.1\%, and typically 1-3\%.

\subsection{Training and inference times}
\label{supp:subsec:training_and_inference_times}

We report training and inference times on LLaVA-1.5-7B. 
For CLIP-UP-Emb, training takes 23.4 minutes on multiple-choice data and 19.7 minutes on open-ended data. 
For CLIP-UP-EmbLoRA, training takes 27.2 minutes for multiple-choice and 22.4 minutes for open-ended data.

CLIP-UP-Emb's impact on inference time is minimal, as the new embedding vector is generated once per input and cached for reuse. 
On the IVQD sub-benchmark from MM-UPD, CLIP-UP-Emb reduces total inference time (15.1 vs. 17.8 minutes) as it generates shorter responses, despite a 16\% increase in per-token time (0.0519 seconds per token vs. 0.0451 seconds).
For CLIP-UP-EmbLoRA, inference time increases due to the LoRA weights not being merged into the base model. In this case, inference takes 21.4 minutes with 0.0748 seconds per-token.

\section{Additional ablation studies}
\label{supp:subsec:additional_ablations}

\cref{table:additional_ablations} presents the full ablation results, including performance for each unanswerability category and additional experiments. 
We examine the effect of the training data by evaluating CLIP-UP-Emb when trained only on data from a single challenge (CLIP-UP-Emb-AAD/IASD/IVQD), but tested on all challenges (lines 1–3 in \cref{table:additional_ablations}). 
As expected, each model performs best on the challenge it was trained on, and outperforms CLIP-UP-Emb (trained on all challenges).
Each such specific model also shows gains, although limited, on the other challenges. 
For example, training on AAD data yields reasonable performance for IASD, a point also observed in~\cite{miyai2025unsolvable}. 
We postulate this is because challenges are interrelated. For instance, IASD may be seen as an extreme case of AAD, where the answer options are not only incorrect but also irrelevant to the question.

\begin{table*}[!t]
    \centering
    \setlength{\tabcolsep}{2.2pt}

        \begin{tabular}{c|ccc|ccc|ccc|ccc}
        \toprule
        \multirow{2}{*}{\textbf{Method}} & \multicolumn{3}{c|}{\textbf{AAD}} & \multicolumn{3}{c|}{\textbf{IASD}} & \multicolumn{3}{c|}{\textbf{IVQD}} & \multicolumn{3}{c}{\textbf{Average}} \\
        & Stand. & UPD & Dual 
        & Stand. & UPD & Dual 
        & Stand. & UPD & Dual
        & Stand. & UPD & Dual \\
        \midrule
        \multicolumn{1}{l|}{CLIP-UP-Emb-AAD} & 
        63.78 & 87.44 & \textbf{59.15} &
        61.48 & 71.16 & 44.07 &
        55.06 & 53.93 & 32.58 & 
        60.11 & 70.84 & 45.27 \\  
        
        \multicolumn{1}{l|}{CLIP-UP-Emb-IASD} & 
        63.41 & 48.66 & 34.76 &
        60.72 & 92.82 & \textbf{56.15} &
        60.11 & 62.92 & 41.29 &
        61.41 & 68.13 & 44.07 \\  
        
        \multicolumn{1}{l|}{CLIP-UP-Emb-IVQD} &
        56.10 & 23.05 & 9.27 &
        53.21 & 51.69 & 25.90 & 
        58.15 & 88.20 & \textbf{51.97} &
        55.82 & 54.31 & 29.05 \\ 
        
        \multicolumn{1}{l|}{CLIP-UP-Emb w\slash{} const. signal} &
        59.27 & 59.39 & 43.05 &
        56.69 & 88.03 & 49.51 &
        54.49 & 78.09 & 43.26 & 
        56.82 & 75.17 & 45.27 \\
        
        \multicolumn{1}{l|}{CLIP-UP-Emb w\slash{} similarities} &
        62.32 & 55.61 & 42.44 &
        59.85 & 90.64 & 54.62 &
        58.99 & 75.28 & 46.63 &
        60.39 & 73.84 & 47.90 \\
        
        \multicolumn{1}{l|}{CLIP-UP-Emb w\slash{} CLIP ViT-L/14} &
        50.73 & 65.12 & 35.00 &
        47.01 & 89.88 & 42.66 &
        46.07 & 87.92 & 39.33 &
        47.94 & 80.97 & 39.00 \\
        
        \multicolumn{1}{l|}{CLIP-UP-Emb (ours)} & 
        61.22 & 67.68 & 47.80 &
        59.52 & 90.64 & 54.73 &
        58.43 & 82.58 & 51.12 &
        59.72 & 80.30 & \textbf{51.22} \\

        \hline
        
        \multicolumn{1}{l|}{CLIP-UP-Emb (const. signal) + LoRA} & 
        61.46 & 58.66 & 37.56 &
        58.11 & 93.47 & 53.54 &
        56.74 & 86.24 & 47.75 &
        58.77 & 79.46 & 46.28 \\
        
        \multicolumn{1}{l|}{CLIP-UP-Emb + LoRA} &
        61.22 & 72.56 & \textbf{49.27} & 
        59.30 & 92.38 & \textbf{55.17} &
        58.99 & 85.11 & \textbf{52.81} & 
        59.84 & 83.35 & \textbf{52.42} \\
        
        \hline
        
        \multicolumn{1}{l|}{CLIP-UP-EmbLoRA (const. signals)} &
        62.93 & 59.88 & 41.10 &
        59.30 & 93.80 & 55.39 &
        58.15 & 88.48 & 50.56 & 
        60.13 & 80.72 & 49.02 \\
        
        \multicolumn{1}{l|}{CLIP-UP-EmbLoRA (ours)} & 
        62.44 & 71.71 & \textbf{49.27} & 
        60.17 & 93.47 & \textbf{56.58} & 
        59.55 & 84.83 & \textbf{53.37} & 
        60.72 & 83.34 & \textbf{53.07} \\
        
        \bottomrule
    \end{tabular}
    \caption{Full ablation results (\%) on LLaVA-1.5-7B. 
    Best dual accuracies for each CLIP-UP setting are bolded.}
    \label{table:additional_ablations}
\end{table*}

We also test the effect of using correlation vectors from Structure-CLIP on CLIP-UP-Emb trained with standard LoRA, and on CLIP-UP-EmbLoRA (lines 8–9 and 10–11).
Consistent with the ablations presented in the main paper, removing the correlation vectors leads to a performance drop.

\begin{table*}[!t]
    \centering
    \setlength{\tabcolsep}{2.2pt}

        \begin{tabular}{c|ccc|ccc|ccc}
        \toprule
        \multirow{2}{*}{\textbf{Category}} & \multicolumn{3}{c|}{\textbf{LLaVA-1.5-7B}} & \multicolumn{3}{c|}{\textbf{InternVL3-8B}} & \multicolumn{3}{c}{\textbf{Ovis2-16B}} \\
        & Dual & TPR & TNR 
        & Dual & TPR & TNR
        & Dual & TPR & TNR \\
        \midrule

        \multicolumn{1}{l|}{Coarse Perception} & 
        75.37 & 93.87 & 94.08 &
        89.08 & 96.51 & 97.50 &
        86.08 & 95.59 & 98.70 \\

        \multicolumn{1}{l|}{Attribute Reasoning} & 
        52.68 & 93.87 & 78.39 &
        79.87 & 95.51 & 94.51 &
        81.88 & 96.53 & 95.60 \\

        \multicolumn{1}{l|}{Fine-grained Perception (Instance-Level)} & 
        55.07 & 90.49 & 88.57 &
        80.18 & 92.68 & 97.48 &
        80.48 & 93.52 & 98.70 \\
        
        \multicolumn{1}{l|}{Fine-grained Perception (Cross-Instance)} & 
        47.75 & 80.56 & 92.54 &
        74.32 & 91.52 & 96.01 &
        82.43 & 93.42 & 98.46 \\
        
        \multicolumn{1}{l|}{Relation Reasoning} & 
        38.67 & 83.50 & 87.98 &
        76.67 & 94.87 & 92.19 &
        79.33 & 93.20 & 99.31 \\

        \multicolumn{1}{l|}{Logic Reasoning} & 
        12.93 & 91.28 & 60.48 &
        47.62 & 88.94 & 87.01 &
        55.78 & 83.40 & 94.25 \\

        \hline
        
        \multicolumn{1}{l|}{All Categories} & 
        53.17 & 89.78 & 86.87 &
        78.71 & 93.90 & 95.47 &
        80.24 & 93.68 & 98.05 \\
        
        \bottomrule
    \end{tabular}
    \caption{Dual accuracy, true positive rate, and true negative rate (all in \%) for LLaVA-1.5-7B, InternVL3-8B, and Ovis2-16B enhanced with CLIP-UP-EmbLoRA, evaluated across different MM-UPD~\cite{miyai2025unsolvable} categories.}
    \label{table:analysis}
\end{table*}

\section{Limitations}
\label{supp:sec:limitations}

Although shown to significantly enhance VLMs' UPD performance, CLIP-UP has several limitations to be acknowledged. It depends on the quality of the CLIP signal, which introduces potential shortcomings since CLIP is known to struggle with issues such as attribute binding and spatial reasoning~\cite{yuksekgonul2023when}. While we mitigate some of these challenges by using Structure-CLIP, others remain.

To analyze these limitations, we report dual accuracy, true positive rate (TPR), and true negative rate (TNR) on different MM-UPD question categories in \cref{table:analysis}, for three VLMs enhanced with CLIP-UP-EmbLoRA.
``Positive'' refers to the VLM choosing to refrain from answering. Thus, TPR measures how often a VLM withholds answers to unanswerable questions (\ie, responds with ``I cannot answer''), while TNR measures how often it chooses to answer answerable questions.

Some categories are more difficult, such as Relation Reasoning and Logical Reasoning, yielding lower dual accuracies. 
These are also categories where Structure-CLIP is expected to struggle, leading to lower TNRs (\ie, over-abstention).
However, stronger models (InternVL3-8B and Ovis2-16B) show more stable TNRs on these categories, suggesting they are less sensitive to such limitations.
\cref{fig:mm_upd_aad_limitation_example} shows an example from the Logical Reasoning category where Structure-CLIP is less indicative and CLIP-UP is limited.

\paragraph{Finer-grained unanswerability}

\begin{table*}[!t]
    \centering
    \begin{tabular}{c|ccc|ccc|ccc}
        \toprule
        \multirow{2}{*}{\textbf{Method}} &
        \multicolumn{3}{c|}{\textbf{LLaVA-1.5-7B}} &
        \multicolumn{3}{c|}{\textbf{LLaVA-NeXT-13B}} &
        \multicolumn{3}{c}{\textbf{Phi-3.5-Vision}} \\
        & Stand. & UPD & Dual &
        Stand. & UPD & Dual & 
        Stand. & UPD & Dual \\
        \midrule
        \multicolumn{1}{l|}{Original Model} & 
        74.40 & 0.00 & 0.00 &
        67.60 & 0.00 & 0.00 &  
        61.20 & 0.00 & 0.00 \\
        
        \multicolumn{1}{l|}{Base Setting} & 
        73.20 & 0.00 & 0.00 &
        69.60 & 1.20 & 1.20 &  
        61.60 & 0.00 & 0.00 \\

        \multicolumn{1}{l|}{Additional-Option} & 
        53.60 & 0.00 & 0.00 &
        68.40 & 0.00 & 0.00 &  
        62.00 & 0.00 & 0.00 \\
        
        \multicolumn{1}{l|}{Additional-Instruction} & 
        70.40 & 2.80 & 1.60 &
        73.20 & 1.20 & 1.20 &  
        62.00 & 8.00 & 6.00 \\
        
        \multicolumn{1}{l|}{LoRA Fine-Tuning} &
        75.60 & 2.40 & 2.40 &
        65.20 & 0.00 & 0.00 &  
        65.20 & 32.40 & 21.20 \\
        
        \multicolumn{1}{l|}{CLIP-UP-Emb (ours)} &
        44.80 & 74.80 & \textbf{31.60} &
        44.00 & 64.40 & \textbf{25.60} &  
        38.00 & 74.80 & \textbf{26.00} \\
        
        \multicolumn{1}{l|}{CLIP-UP-EmbLoRA (ours)} &
        46.00 & 70.40 & \textbf{31.60} &
        59.20 & 7.20 & 3.60 & 
        49.60 & 54.80 & 23.20 \\

        \bottomrule
    \end{tabular}
    \vfill
    \vspace{0.3cm}
    \begin{tabular}{c|ccc|ccc|ccc}
        \toprule
        \multirow{2}{*}{\textbf{Method}} &
        \multicolumn{3}{c|}{\textbf{InternVL3-1B}} &
        \multicolumn{3}{c|}{\textbf{InternVL3-8B}} &
        \multicolumn{3}{c}{\textbf{Ovis2-16B}} \\
        & Stand. & UPD & Dual 
        & Stand. & UPD & Dual 
        & Stand. & UPD & Dual \\
        \midrule
        \multicolumn{1}{l|}{Original Model} & 
        70.00 & 0.00 & 0.00 & 
        80.80 & 0.00 & 0.00 & 
        76.40 & 0.00 & 0.00\\

        \multicolumn{1}{l|}{Base Setting} & 
        71.60 & 0.00 & 0.00 &
        78.80 & 0.00 & 0.00 &
        76.40 & 5.60 & 3.20 \\

        \multicolumn{1}{l|}{Additional-Option} & 
        72.00 & 0.00 & 0.00 &
        82.00 & 0.00 & 0.00 &
        76.80 &  0.00 & 0.00 \\

        \multicolumn{1}{l|}{Additional-Instruction} & 
        72.80 & 0.00 & 0.00 &
        79.20 & 0.00 & 0.00 &
        76.00 & 16.00 & 11.60 \\

        \multicolumn{1}{l|}{LoRA Fine-Tuning} &
        70.00 & 2.00 & 1.60 &
        79.20 & 1.20 & 1.20 &
        78.80 & 3.60 & 2.80 \\

        \multicolumn{1}{l|}{CLIP-UP-Emb (ours)} &
        54.80 & 30.00 & 17.60 &
        79.20 & 27.20 & \textbf{20.00} &
        72.40 & 52.80 & \textbf{38.40} \\

        \multicolumn{1}{l|}{CLIP-UP-EmbLoRA (ours)} &
        59.60 & 46.40 & \textbf{26.80} &
        76.40 & 5.20 & 4.40 &
        74.80 & 8.40 & 5.20 \\

        \bottomrule
    \end{tabular}

    \caption{
    Results (\%) on UVQA~\cite{he2024tubenchbenchmarkinglargevisionlanguage} multiple-choice VQA subset.
    Metrics include standard, UPD, and dual accuracies.}
    \label{table:tubench_uvqa_results}
\end{table*}

\begin{table*}[!t]
    \centering
    \begin{tabular}{c|ccc|ccc}
        \toprule
        \multirow{2}{*}{\textbf{Method}} & \multicolumn{3}{c|}{\textbf{UCR}} & \multicolumn{3}{c}{\textbf{UTabMWP}} \\ 
        & Stand. & UPD & Dual & Stand. & UPD & Dual \\
        \midrule
        \multicolumn{1}{l|}{Original Model} & 
        49.07 & 0.00 & 0.00 & 
        49.50 & 0.00 & 0.00 \\
        
        \multicolumn{1}{l|}{Base Setting} & 
        50.93 & 0.00 & 0.00 & 
        44.50 & 1.00 & 0.50 \\ 
        
        \multicolumn{1}{l|}{Additional-Option} & 
        52.78 & 0.00 & 0.00 & 
        44.00 & 0.00 & 0.00 \\ 

        \multicolumn{1}{l|}{Additional-Instruction} & 
        47.22 & 0.00 & 0.00 & 
        46.00 & 6.50 & 0.50 \\ 

        \multicolumn{1}{l|}{LoRA Fine-Tuning} & 
        50.93 & 0.00 & 0.00 & 
        40.00 & 22.00 & 1.00 \\ 

        \multicolumn{1}{l|}{CLIP-UP-Emb (ours)} &
        35.19 & 20.37 & 9.26 & 
        17.50 & 43.00 & 0.00 \\ 

        \multicolumn{1}{l|}{CLIP-UP-EmbLoRA (ours)} &
        22.22 & 40.74 & 7.41 & 
        26.00 & 35.50 & 0.50 \\ 

        \bottomrule
    \end{tabular}
    \caption{
    Results (\%) on the UCR and UTabMWP~\cite{he2024tubenchbenchmarkinglargevisionlanguage} multiple-choice VQA subsets for LLaVA-1.5-7B.
    Metrics include standard, UPD, and dual accuracies.}
    \label{table:tu_bench_ucr_utabmwp_results}
\end{table*}

To investigate this point further, we evaluate CLIP-UP on three subsets from the multiple-choice TUBench benchmark~\cite{he2024tubenchbenchmarkinglargevisionlanguage}: UVQA, UCR, and UTabMWP.
Unanswerable questions in TUBench were created by applying fine-grained edits to the text, keeping the questions grounded in the image but without a correct answer, with some unanswerable due to missing or indeterminate information.
Specifically, UVQA includes 250 answerable-unanswerable question pairs about natural images that require nuanced reasoning for answerability assessment.

\cref{table:tubench_uvqa_results} presents the results on the UVQA subset.
The original models and prompt engineering baselines exhibit low abstention rates, indicating that detecting unanswerable questions is particularly difficult on this benchmark.
LoRA fine-tuning mostly has minimal effect on model behavior, while CLIP-UP variants generally improve dual accuracy. 
However, the gains are smaller, less consistent, and often come at the cost of a steep drop in standard accuracy.

This is somewhat expected, given the fine-grained nature of UVQA answerability and the questions being in a yes/no format (converted to multiple-choice). Still, CLIP-UP produces gains, for example, standard drops in CLIP-UP on InternVL3-8B and Ovis2-16B are reasonable.

We additionally test on two harder subsets from TUBench: 480 yes/no question pairs from the UCR subset focused on code snippets, and 108 multiple-choice question pairs from the UTabMWP subset focused on tabular data.  
\cref{table:tu_bench_ucr_utabmwp_results} shows the results on LLaVA-1.5-7B. Although CLIP-UP generally outperforms other methods, the gains are smaller.

\paragraph{Further discussion}

However present, these limitations are not inherent to CLIP-UP itself, as Structure-CLIP can be replaced with another CLIP variant or even a non-CLIP model that produces shared vision-language embeddings.
Thus, given an alignment model that provides a suitable signal, CLIP-UP could potentially support better handling of more fine-grained types of unanswerability.

Finally, a more fundamental limitation relates to the reliance on image-text alignment for open-ended questions. 
For questions with very general prompts (\eg, ``What do you see in the image?''), the alignment signal may not distinguish between answerable and unanswerable cases, making CLIP-UP less effective in such scenarios.

\begin{figure*}[!t]
\small
\centering
\begin{tabular}{L{0.16\textwidth}|L{0.37\textwidth}|L{0.37\textwidth}}
\toprule
 & \multicolumn{1}{c|}{\textbf{Standard}} & \multicolumn{1}{c}{\textbf{Unanswerable}} \\
\midrule

\multirow{2}{=}{\makecell{\textbf{Question}}} 
& \multicolumn{2}{c}{\makecell[l]{%
Based on the description, how are the people in the image engaging with the game?\\
A. The group of people is physically engaging with the game by using Nintendo Wii controllers.\\
B. The group of people is physically engaging with the game by using traditional gaming controllers.\\
C. The group of people is engaging with the game by watching a screen passively.\\
D. The group of people is engaging with the game by playing a board game.\\
}} \\

& \makebox[0.9\linewidth]{\includegraphics[height=0.16\textwidth]{images/predictions_demo/image_standard_ivqd_205.png}}
& \makebox[0.9\linewidth]{\includegraphics[height=0.16\textwidth]{images/predictions_demo/image_upd_ivqd_205.png}} \\
\midrule

LLaVA-1.5-7B 
& A. The group of people is physically engaging with the game by using Nintendo Wii controllers. \cmark 
& C. The group of people is engaging with the game by watching a screen passively. \xmark \\
\graymidrule
LLaVA-1.5-7B + CLIP-UP-EmbLoRA
& A. The group of people is physically engaging with the game by using Nintendo Wii controllers. \cmark & 
I cannot answer. \cmark \\
\midrule

LLaVA-NeXT-13B
& A. The group of people is physically engaging with the game by using Nintendo Wii controllers. \cmark 
& A. The group of people is physically engaging with the game by using Nintendo Wii controllers. \xmark \\
\graymidrule
LLaVA-NeXT-13B + CLIP-UP-EmbLoRA
& A. The group of people is physically engaging with the game by using Nintendo Wii controllers \cmark & 
I cannot answer. \cmark \\
\midrule

Phi-3.5-Vision
& A. \cmark 
& A. \xmark \\
\graymidrule
Phi-3.5-Vision + CLIP-UP-EmbLoRA
& A. \cmark & 
I cannot answer. \cmark \\
\midrule

InternVL3-1B
& A. The group of people is physically engaging with the game by using Nintendo Wii controllers.
In the image, the individuals are actively participating in a video game using Nintendo Wii controllers. The person on the right is holding a controller, and the person in the middle is also holding a controller, indicating that they are actively engaged in the game. The other individuals are standing around, watching the game, but they are not using controllers; instead, they are observing the screen. \cmark 
& C. The group of people is engaging with the game by watching a screen passively.
The image shows a group of people watching a screen passively, which suggests they are not actively participating in the game but rather observing it. This is a common scenario in gaming settings where the focus is on the content displayed on the screen rather than the gameplay itself. \xmark \\
\graymidrule
InternVL3-1B + CLIP-UP-EmbLoRA
& A. The group of people is physically engaging with the game by using Nintendo Wii controllers. \cmark & 
I cannot answer. \cmark \\
\midrule

InternVL3-8B
& A. The group of people is physically engaging with the game by using Nintendo Wii controllers. \cmark 
& A. The group of people is physically engaging with the game by using Nintendo Wii controllers. \xmark \\
\graymidrule
InternVL3-8B + CLIP-UP-EmbLoRA
& A. The group of people is physically engaging with the game by using Nintendo Wii controllers. \cmark & 
I cannot answer. \cmark \\
\midrule

Ovis2-16B
& The description mentions that the people are playing a game on a Nintendo Wii, which involves physical interaction with the game using motion-sensitive controllers. Therefore, the correct answer is A. \cmark 
& The description mentions that the group of people is physically engaging with the game, using Nintendo Wii controllers, which suggests an active and interactive form of play. Therefore, the correct answer is A. \xmark \\
\graymidrule
Ovis2-16B + CLIP-UP-EmbLoRA
& A. The group of people is physically engaging with the game by using Nintendo Wii controllers. \cmark & 
I cannot answer. \cmark \\
\midrule
\end{tabular}
\caption{Models predictions on a pair of standard and IVQD unanswerable multiple-choice VQA questions from MM-UPD~\cite{miyai2025unsolvable}, without and with CLIP-UP-EmbLoRA.}
\label{fig:mm_upd_ivqd_performance_example}
\end{figure*}

\begin{figure*}[!t]
\small
\centering
\begin{tabular}{L{0.16\textwidth}|L{0.37\textwidth}|L{0.37\textwidth}}
\toprule
 & \multicolumn{1}{c|}{\textbf{Standard}} & \multicolumn{1}{c}{\textbf{Unanswerable}} \\
\midrule

\multirow{3}{=}{\makecell{\textbf{Question}}} 
& \multicolumn{2}{c}{%
    \includegraphics[width=0.22\textwidth]{images/predictions_demo/image_upd_rgqa_31626.png}
} \\
& \parbox[t]{0.37\textwidth}{Where are the clothes?} 
& \parbox[t]{0.37\textwidth}{Where are the pink clothes?}  \\
\midrule

LLaVA-1.5-7B 
& Suitcase. \cmark 
& Suitcase. \xmark \\
\graymidrule
\makecell[l]{LLaVA-1.5-7B +\\CLIP-UP-EmbLoRA} 
& Suitcase. \cmark 
& I cannot answer. \cmark \\
\midrule

LLaVA-NeXT-13B
& Suitcase. \cmark 
& Suitcase. \xmark \\
\graymidrule
LLaVA-NeXT-13B + CLIP-UP-EmbLoRA
& Suitcase. \cmark 
& I cannot answer. \cmark \\
\midrule

Phi-3.5-Vision
& Suitcase. \cmark 
& Suitcase. \xmark \\
\graymidrule
Phi-3.5-Vision + CLIP-UP-EmbLoRA
& Suitcase. \cmark 
& I cannot answer. \cmark \\
\midrule

InternVL3-1B
& Suitcase. \cmark 
& Bag. \xmark \\
\graymidrule
InternVL3-1B + CLIP-UP-EmbLoRA
& Bag. \cmark 
& I cannot answer. \cmark \\
\midrule

InternVL3-8B
& Suitcase. \cmark 
& Suitcase. \xmark \\
\graymidrule
InternVL3-8B + CLIP-UP-EmbLoRA
& Suitcase. \cmark 
& I cannot answer. \cmark \\
\midrule

Ovis2-16B
& Suitcase. \cmark 
& Suitcase. \xmark \\
\graymidrule
Ovis2-16B + CLIP-UP-EmbLoRA
& Suitcase. \cmark 
& I cannot answer. \cmark \\
\midrule

\end{tabular}
\caption{Models predictions on a pair of standard and unanswerable open-ended VQA questions from RGQA~\cite{zhang2023toward}, without and with CLIP-UP-EmbLoRA.}
\label{fig:rgqa_performance_example}
\end{figure*}

\begin{figure*}[!t]
\small
\centering
\begin{tabular}{L{0.16\textwidth}|L{0.37\textwidth}|L{0.37\textwidth}}
\toprule
 & \multicolumn{1}{c|}{\textbf{Standard}} & \multicolumn{1}{c}{\textbf{Unanswerable}} \\
\midrule

\multirow{3}{=}{\makecell{\textbf{Question}}} 
& \multicolumn{2}{c}{
    \includegraphics[width=0.215\textwidth]{images/predictions_demo/image_upd_aad_459.png}
} \\
& \parbox[t]{0.37\textwidth}{Which letter marks the territory controlled by the\\ancient Maya civilization?\\A. B\\B. C\\C. A\\D. D} 
& \parbox[t]{0.37\textwidth}{Which letter marks the territory controlled by the\\ancient Maya civilization?\\A. B\\B. A\\C. D} \\
\midrule

LLaVA-1.5-7B 
& A. \xmark 
& A. \xmark \\
\graymidrule
LLaVA-1.5-7B + CLIP-UP-EmbLoRA
& A. B. \xmark 
& A. B. \xmark \\
\midrule

LLaVA-NeXT-13B
& B. \cmark 
& C. \xmark \\
\graymidrule
LLaVA-NeXT-13B + CLIP-UP-EmbLoRA
& D.D. \xmark 
& C.D. \xmark \\
\midrule

Phi-3.5-Vision
& B. \cmark 
& B. \xmark \\
\graymidrule
Phi-3.5-Vision + CLIP-UP-EmbLoRA
& B. \cmark 
& B. \xmark \\
\midrule

InternVL3-1B
& To determine which letter marks the territory controlled by the ancient Maya civilization... Given this information, the correct answer is: B. \cmark
& To determine which letter marks the territory controlled by the ancient Maya civilization... Given this information, the correct answer is: A. \xmark \\
\graymidrule
InternVL3-1B + CLIP-UP-EmbLoRA
& B. C. \cmark
& I cannot answer. \cmark \\
\midrule

InternVL3-8B
& To determine which letter marks the territory controlled by the ancient Maya civilization... Therefore, the letter that marks the territory controlled by the ancient Maya civilization is D. \xmark 
& To determine which letter marks the territory controlled by the ancient Maya civilization... Therefore, the letter that marks the territory controlled by the ancient Maya civilization is D. \xmark \\
\graymidrule
InternVL3-8B + CLIP-UP-EmbLoRA
& B. C. \cmark 
& I cannot answer. \cmark \\
\midrule

Ovis2-16B
& The ancient Maya civilization was located in the southeastern part of Mesoamerica, which corresponds to the region marked by letter B on the map. This area includes parts of present-day Mexico, Guatemala, Belize, Honduras, and El Salvador. The Maya civilization was known for its advanced writing system, architecture, and astronomical knowledge. Therefore, the correct answer is A. B. \xmark 
& The letter that marks the territory controlled by the ancient Maya civilization is B. The Maya civilization was centered in the region that is now southeastern Mexico, including the Yucatán Peninsula, and parts of Guatemala, Belize, Honduras, and El Salvador. This area corresponds to the region marked by letter B on the map. \xmark \\
\graymidrule
Ovis2-16B + CLIP-UP-EmbLoRA
& D. D. \xmark 
& C. D. \xmark \\
\midrule

\end{tabular}
\caption{Model predictions on standard and unanswerable multiple-choice VQA questions from MM-UPD~\cite{miyai2025unsolvable}, illustrating the limitations of CLIP-UP-EmbLoRA.}
\label{fig:mm_upd_aad_limitation_example}
\end{figure*}